\documentclass[12pt,oneside,a4paper,notitlepage]{article}
\usepackage{array}
\usepackage{graphicx}
\usepackage[format=plain,
            labelfont={bf,it},
            textfont=it]{caption}
\usepackage{wrapfig}
\usepackage{amsmath,amssymb}
\usepackage[polish,english]{babel}
\usepackage{graphicx}
\usepackage{url}
\usepackage{perpage}
\MakePerPage{footnote}
\usepackage{enumitem}
\setlist{nolistsep}
\graphicspath{{figs/}}
\newdimen\plWleft
\newdimen\plWdown
\newdimen\plWright
\newdimen\plWtemp
\def\sob#1#2#3#4#5{%parameters: letter and fractions hl,ho,vl,vo
  \setbox0\hbox{#1}\setbox1\hbox{$_\mathchar'454$}\setbox2\hbox{p}%
  \plWright=#2\wd0 \advance\plWright by-#3\wd1
  \plWdown=#5\ht1 \advance\plWdown by-#4\ht0
  \plWleft=\plWright \advance\plWleft by\wd1
  \plWtemp=-\plWdown \advance\plWtemp by\dp2 \dp1=\plWtemp
  \leavevmode
  \kern\plWright\lower\plWdown\box1\kern-\plWleft #1}
\DeclareTextCommand{\eob}{OT1}{\sob e{.50}{.35}{0}{.93}}
\DeclareTextCommand{\Eob}{OT1}{\sob E{.60}{.35}{0}{.90}}
\DeclareTextCommand{\eob}{T1}{\k e} \DeclareTextCommand{\aob}{OT1}{\sob
a{.66}{.20}{0}{.90}} \DeclareTextCommand{\aob}{T1}{\k a}

\title{Fast kNN mode seeking clustering \\applied to active learning}

\author{Robert P.W. Duin \\
PRLab, Delft University of Technology, Netherlands \\r.p.w.duin@tudelft.nl
\and Sergey Verzakov \\ Prime Vision, Delft, Netherlands \\s.verzakov@gmail.com}

\begin{document}

\maketitle

\begin{abstract}
\noindent
A significantly faster algorithm is presented for the original kNN mode seeking procedure.
It has the advantages over the well-known mean shift algorithm that it is feasible in high-dimensional vector spaces and results in uniquely, well defined modes. Moreover, without any additional computational effort it may yield a multi-scale hierarchy of clusterings. The time complexity is just $O(n \sqrt n)$. Resulting computing times range from seconds for $10^4$ objects to minutes for $10^5$ objects and to less than an hour for $10^6$ objects. The space complexity is just $O(n)$. The procedure is well suited for finding large sets of small clusters and is thereby a candidate to analyze thousands of clusters in millions of objects.

The kNN mode seeking procedure can be used for active learning by assigning the clusters to the class of the modal objects of the clusters. Its feasibility is shown by some examples with up to 1.5 million handwritten digits. The obtained classification results based on the clusterings are compared with those obtained by the nearest neighbor rule and the support vector classifier based on the same labeled objects for training. It can be concluded that using the clustering structure for classification can be significantly better than using the trained classifiers. A drawback of using the clustering for classification, however, is that no classifier is obtained that may be used for out-of-sample objects.

\end{abstract}

\section{Introduction}
\label{section:intr}
Mode seeking clustering is based on finding the modes of the estimated probability density function of a given set of objects. For every mode a cluster is defined consisting of all objects for which the density gradient followed from that object arrives at the particular mode. The basic idea can be traced back to two papers by Fukunaga et al. \cite{Fukunaga75}, \cite{Koontz76} in the seventies. In order to be able to handle arbitrarily shaped clusters non-parametric density estimates are needed. Fukunaga et al. considered the two well known ways: Parzen kernel densities as well as estimates based on the $k$-Nearest Neighbors (kNN).

The kernel density estimate has been studied extensively for mode seeking clustering. It has been made most popular by the mean shift algorithm \cite{Comaniciu02}. This is based on gradient estimates computed from the derivative of the kernel function. For higher dimensionalities iteratively computing and following the gradient is a computationally intensive procedure. Moreover, the number of restarts that is needed is a multiple of the number of modes. A randomly selected subset of objects is used for starting the gradient searches.

The mean shift procedure is particulary useful for low dimensional datasets with a small or moderate number of clusters. An additional problem to be solved is the selection of the kernel. A small one will result in many clusters, a large one will combine them and may yield a too smooth result. In practise this may be solved by running the entire procedure for a set of kernels.

The kNN mode seeking clustering as described in \cite{Duin2012} is simpler, faster, but less accurate than the original proposal by Koontz and Fukunaga \cite{Koontz76}. Densities are not exactly computed, but are just related to the distance to the $k$-th nearest neighbor. The smaller $k$, the higher the clustering resolution and the more modes are found. Gradient steps are defined by setting and following pointers from every object to the one with the highest density in its neighborhood, which we will call the \emph{modal object}\footnote{The concept of modal objects has been used in a different way outside statistical data analysis in the field of natural languages by F. Moltmann, see, A Predicativist Semantics of Modals Based on Modal Objects, Proc. of the Amsterdam Colloquium 2015.}. Consequently, after computing all $n \times n$ distances between the $n$ objects in a dataset all modes can be found by following the pointers from all objects. The computational effort is negligible compared to the distance computations of the previous step. In \cite{Duin2012} it is discussed how to avoid the storage of all distances.

A big advantage of the proposal is that different clustering resolutions related to a set of neighborhood sizes $k$ can be simultaneously computed without the need of any recomputation. This can be achieved by just the cost of storing multiple pointers, which is feasible as the full distance matrix is not stored. The total space complexity is $O(nd +nm)$ in which $d$ is the number of features and $m$ is the number of neighborhood sizes.

From the application point of view, kNN mode seeking has the advantage over the mean shift algorithm that an automatic scaling is included in a $k$-nearest neighbor approach. Objects in less dense areas will be related to more distant objects than in high density regions. In the mean shift algorithm kernels have the same shape over the entire vector space.

In this paper a fast procedure for kNN mode seeking is presented, evaluated and applied to the labeling of large datasets. In Section \ref{section:algorithm} the fast algorithm is discussed and it is shown that it approximates the same clustering by computing just $n \times \sqrt n$ instead of all $n \times n$ distances. This is achieved by constructing a set of overlapping cells by which for every object the number of candidates that might belong to the $k$ nearest neighbors is significantly reduced. Whether this approximation is appropriate depends on the structure of the data and the type of clusters to be searched and thereby on the application.

In Section \ref{sec:classification} various possibilities of the labeling of a large dataset are considered based on mode seeking clustering. By using the labels of just the modal objects the clustering can be used for labeling all other objects, resulting in an active labeling procedure. This classification scheme can be improved by using the multi-level clustering property of the kNN mode seeking procedure that is obtained by almost no additional computational effort. The same structure can be used for including a reject option.

The proposed procedures are illustrated and evaluated is Section \ref{section:Hand-written digits} by some of large datasets of handwritten characters ranging from 70000 objects to almost 1,5 million objects. Computing times, cluster qualities (compared to true labels) and classification performances as a function of the number of labeled objects (learning curves) are measured and graphically shown. Conclusions are summarized in the final Section \ref{section:Conclusions}. The software and some dataset used in this paper are publicly available \cite{Duin_ct}.

\section{Fast kNN mode seeking algorithm}
\label{section:algorithm}

\subsection{The original kNN mode seeking algorithm}
\label{section:old algorithm}
In the version of kNN mode seeking, as described in \cite{Duin2012}, object densities are defined as the reverse of the distance to the $k$-th nearest neighbor. The below algorithm is used. It is good for very large datasets with $n > 10^5$ objects. All pairwise distances are needed two times. As $n^2 = 10^{10}$ distances cannot be stored, they are computed twice. This is done for a set of $m$ neighborhood sizes $k$ (e.g. $m=25$) in parallel, by which $nm$ densities and $nm$ pointers have to be stored. These are used to compute $nm$ cluster indices for $m$ different clustering resolutions of the $n$ objects.

\vspace{5mm}
\begin{enumerate}
  \item S is the user supplied set of objects with size $n$.
  \item $K$ is a user defined set of target neighborhood sizes $K = \{k_1,k_2, ...k_m\}$.
  \item Repeat for all $n$ objects $\textbf{x}_i$ in S ($i=1,2, \ldots ,n$):
  \item ~~Compute its distances $d_{ij}$ to all other objects $\textbf{x}_j$ in S.
  \item ~~Sort them: $\textbf{s}_i = sort_j(d_{ij})$
  \item ~~Store density estimates $\forall k \in K: f_i^k = 1/s_{ik}$.
  \item Next $i$
  \item Repeat for all $n$ objects $\textbf{x}_i$ in S ($i=1,2, \ldots ,n$):
  \item ~~Compute its distances $d_{ij}$ to all other objects $\textbf{x}_j$ in S.
  \item ~~Rank them: $t_i = argsort_j(d_{ij})$
  \item ~~Store for all $k \in K$ a pointer $u_i^k = argmax_{r \in t_{i,j = 1 \ldots k}}(f_r^k)$
  \item Next $i$
  \item Repeat for all neighborhood sizes $k \in K$
  \item ~~Repeat until no change $\forall i: u_i^k = u_{u_i^k}^k$
  \item ~~Store clustering for neighborhood size $k$: $C_k = [u_1^k,u_2^k, ... u_n^k]$
  \item Next $k$
\end{enumerate}

\vspace{5mm}

In the steps 2-7, the densities are computed for a set of resolutions defined by K, and stored for all objects. The neighborhoods themselves are not stored in this algorithm.  They are recomputed in the next loop 8-12 in order to find, for every neighborhood size K, the objects with the highest densities. The pointers to objects with higher densities are, for all resolutions, followed in the iteration loop 13-16. This final loop takes almost no time in comparison with the two other ones. In case the studied neighborhood sizes are small, e.g. up to 100, the indices of the neighborhood objects can be stored in the first loop. For that case, the computing time of the second loop becomes negligible too. As we are interested in larger neighborhood sizes this does not hold for the experiments described in Section \ref{section:Hand-written digits}.

The time complexity of this algorithm is at least $\textsl{O}(n^2)$ as in the first loop the distances of all objects to all other ones have to be computed. As they are not stored no use of symmetry can be made. Various proposals have been made to speed up the computation of the nearest neighbors, e.g. by the use of k-d trees. They are especially effective in low-dimensional spaces and for an (approximative) determination of the first nearest neighbor. They do not offer a solution for finding the neighbors in a large neighborhood, which is needed to find larger clusters.

In the remainder of this paper we will refer to this original kNN Mode Seeking algorithm by MS.

\subsection{The proposed fast kNN mode seeking algorithm}
\label{section:fast algorithm}
The original algorithm presented in Section \ref{section:old algorithm} is slow for very large datasets due to its time complexity. Every object is compared with every other object. For measuring densities, however, it is sufficient to relate objects just with their neighboring ones. If candidate neighborhood objects could be selected without inspecting them all, a speedup would be realized. In \cite{Dong2011} a procedure is presented which experimentally appeared to have a time complexity of $\textsl{O}(n^{1.14})$. This procedure, however, is based on a fixed neighborhood size, while the advantage of our procedure is that it can handle simultaneously a set of neighborhood sizes. Inspired by the result of \cite{Dong2011} we developed the following procedure.

First a small set P of $m << n$ objects is selected, e.g. at random. It is not essential that P is a subset of the given dataset S, but its feature representation should be in the same domain as S. Next the 1-nearest neighbor graph is computed in which every object in S points to its nearest neighbor in P. In this way S is split into $m$ subsets and the space is split into $m$ cells. For this operation $n m$ distance computations are needed. Cells contain on the average $n / m$ objects but may have different sizes. We name this set of non-overlapping cells the P-cells.

For sufficiently small values of $k$ ($k << n/m$) it is expected that the $k$-nearest neighbors of an object are in the same cell. Consequently just $n (n / m)$ instead of $n^2$ distances need to be computed in addition to find for every object in S its $k$ nearest neighbors. In total this demands thereby at least $n m + n (n / m)$ distance computations (somewhat more if some cells contain more objects than the average, which is to be expected). This is minimum if $m = \sqrt{n}$ is chosen. The total number of distances that has to be computed is thereby at least $2 n \sqrt{n}$.

In every P-cell some objects may be close to the cell border and it might thereby happen that some of their $k$ nearest neighbors in the total dataset S are in a neighboring cell and are thereby not considered. In order to reduce the probability that this happens we enlarged the set of candidates defined by all objects that share the same object in P as their 1-nearest neighbor in P and change it to all objects that share the same object in P as belonging to their set of $c$-nearest neighbors. This results in a set of overlapping cells, to be called Q-cells, that contain in the order of $cn/m$ objects. For every P-cell there is a larger Q-cell that entirely contains the P-cell.

The neighborhood search in the fast mode seeking procedure will, for every object, be restricted to the objects in the Q-cell defined by the P-cell to which the object belongs. If all Q-cells would contain exactly this number of objects the number of distance computations is now minimum for $m = \sqrt{cn}$, resulting in at least $2n \sqrt{cn}$ distance computations. So the time complexity of the fast algorithm is $\textsl{O}(n \sqrt{n})$. The speed-up of the fast algorithm is thereby in the order of $\sqrt{n/c}$.

A formal description of the algorithm is given below. It follows the same three loops as the algorithm in Section \ref{section:old algorithm}. The first two loops, however, are now restricted to the objects in the Q-cells instead of all objects. The P set is chosen at random  from the available dataset S.

\vspace{5mm}

\begin{enumerate}
  \item S is the user supplied set of objects with size $n$.
  \item $c$ is the user defined complexity parameter.
  \item $K$ is a user defined set of target neighborhood sizes $K = \{k_1,k_2, ... k_m\}$.
  \item Select at random a subset P of $m = \sqrt{c n}$ objects out of S.
  \item Determine for every object in S its $c$ nearest neighbors in P.
  \item Construct a set of $m$ P-cells of objects in S that have the same object in P as their nearest neighbor in P.
  \item Construct a set of $m$ Q-cells of objects in S that share the same object in P in their $c$ nearest neighbors in P.
  \item Repeat for all $n$ objects $\textbf{x}_i$ in S ($i=1,2, \ldots ,n$):
  \item ~~Determine the P-cell $p_i$ to which $\textbf{x}_i$ belongs.
  \item ~~Determine the Q-cell $q_i$ to  which $p_i$ belongs.
  \item ~~Compute all distances $d_{ij}$ of $\textbf{x}_i$ to all objects $\textbf{x}_j$ in $q_i$.
  \item ~~Sort them: $\textbf{s}_i = sort_j(d_{ij})$
  \item ~~Store density estimates $\forall k \in K: f_i^k = 1/s_{ik}$.
  \item Next $i$
  \item Repeat for all $n$ objects $\textbf{x}_i$ in S ($i=1,2, \ldots ,n$):
  \item ~~Determine the P-cell $p_i$ to which $\textbf{x}_i$ belongs.
  \item ~~Determine the Q-cell $q_i$ to  which $p_i$ belongs.
  \item ~~Compute all distances $d_{ij}$ of $\textbf{x}_i$ to all objects $\textbf{x}_j$ in $q_i$.
  \item ~~Rank them: $t_i = argsort_{j \in q_i}(d_{ij})$
  \item ~~Store for all $k \in K$ a pointer $u_i^k = argmax_{r \in t_{i,j = 1 \ldots k}}(f_r^k)$
  \item Next $i$
  \item Repeat for all sizes $k \in K$
  \item ~~Repeat until no change $\forall i: u_i^k = u_{u_i^k}$
  \item ~~Store clustering for neighborhood size $k$: $C_k = [u_1^k,u_2^k, ... u_n^k]$
  \item Next $k$
\end{enumerate}

\vspace{5mm}
Our implementation is for the selection of the subset P slightly more complicated. A fully random choice may result is some very small P-cells. This may result in a bad quality of the kNN search. For that reason we remove the few P-cells that contain less than $n/(3m)$ objects. In Section \ref{section:Hand-written digits} some real world examples are given comparing the fast algorithm with the original one.

In the remainder of this paper we will refer to the fast kNN Mode Seeking by FMS or FMS-$c$, in which $c$ is the value of the complexity parameter. By default $c=6$.

\section{Classifying objects based on kNN mode seeking clustering}
\label{sec:classification}
In this section some ways are described to use kNN mode seeking clustering for the classification of a large set of unlabeled objects using a small set of labeled ones. The clustering procedure is applied to the entire dataset. In case the objects to be labeled are determined by the clustering this results in active learning.

\subsection{kNN mode seeking applied to active learning}
\label{section:active learning}

An important application of kNN mode seeking is based on the property that clusters are represented by a single object, the modal object. This is the object with the highest estimated density in the corresponding cluster. All other objects in the cluster have a pointer to this one. Consequently a fast way of labeling the entire dataset is achieved by labeling the modal objects by a human expert, followed by assigning all objects to the class of the modal object of their cluster. This may reduce the cost of labeling considerably. Moreover, it is expected that the modal objects represent the dataset well. Thereby the obtained classification may be more accurate than a classifier trained by a randomly selected training set of the same size. The following procedure is used:

\vspace{5mm}
\begin{enumerate}
  \item Perform MS or FMS clustering on the entire dataset using a set of values for $k$, e.g. $k = 2,3,5,7,10,14,20,30,50,70$, resulting in 10 different clusterings $C_i, i = 1,2, \ldots, 10$.
  \item  Determine the number of clusters $N_i$ for every clustering $C_i$.
  \item  Select an appropriate clustering $C_{i^*}$ w.r.t. the permissable cost of labeling the $N_i$ modal objects found for clustering $C_i$, $i = 1,2, \ldots, 10$.
  \item Label the $N_{i^*}$ modal objects of clustering $C_{i^*}$.
  \item Assign for every cluster in clustering $C_{i^*}$ all objects to the class of their modal object.
\end{enumerate}

\vspace{5mm}
This can be considered as a way of active learning. Therefore we name the procedure MS-AL or FMS-AL. There are, however, some differences with the traditional active learning procedures as the above algorithm is a one-step procedure. There is no intermediate or final classifier computed that is used to select additional objects to be labeled. It seems thereby to be most appropriate for labeling a given dataset as all objects to be labeled are used for estimating the density of these objects. There is no prediction of densities of future, not yet seen objects.

In the above algorithm several clusterings are found, but just a single one is used: the one with an appropriate number of clusters. One may wonder whether the other available clusterings cannot be used for improving the classification. In the next subsections some possibilities are discussed.

\vspace{5mm}
\subsection{kNN mode seeking combined with multi-level \\ confidence based classification}
\label{section:multi-level_clustering}
In the active learning classification procedure as defined above, all objects within a cluster receive the same class assignments. The rationale behind it is that a cluster is expected to contain similar objects. In case more objects in a cluster have known (or estimated) labels or class confidences $\textbf{q}(\textbf{x}_t)$, assigning a cluster average of class confidences to all objects in the cluster follows the same idea:
\begin{equation}
\label{eq:updaterule}
   \textbf{q}_{ij}(\textbf{x}_t) = \frac{1}{|C_{ij}|} \sum_{\textbf{x}_t \in C_{ij}}{\textbf{q}(\textbf{x}_t)}
\end{equation}
Here $\textbf{q}_{ij}(\textbf{x}_t)$ is an estimated row vector of class confidences based on cluster $C_{ij}$, being cluster $j$ of clustering $i$. Let $Q^i = [\textbf{q}_{i1}; \dots ; \textbf{q}_{ir}]$ be the $n \times r$ matrix of the class confidences for the $r$ classes of all $n$ objects and let $A^i$ be an $n \times n$ matrix with
\begin{flalign}
\label{eq:definitionA}
    A^i_{t,t'} &= 1/|C_{ij}| \text{ if } x_t \in C_{ij} \wedge x_{t'} \in C_{ij} \\
               &= 0 \text{ otherwise}
\end{flalign}
We ranked the clusterings from low resolution to high resolution clusterings. The confidences $Q^{i+1}$ are estimated from the confidences $Q^i$ of clustering level $i$ with less clusters by averaging the confidences of level $i$ according to the clustering of level $i+1$:
\begin{equation}
\label{eq:proprule}
    Q^{i+1} = A^{i+1} Q^i
\end{equation}
By iterating over all clustering levels confidence estimates can be obtained for all of them from an initial estimate. This can be obtained from the first level $Q^1$ by active learning. On this level for every cluster the modal object is selected and labeled. All objects in a cluster are given the same confidence: one for the class of the label and zero for all other classes.

The described procedure of propagating and averaging confidences is consistent with the active learning procedure in which all objects receive the same class as the modal object. As on higher clustering levels some clusters may contain objects of different lower resolution clusters the object confidences may receive contributions of various classes.

%We played around with various orderings of the levels and found ranking them from low to high resolution starting with the level used for active learning labeling performed best. This is used in the experiments reported in Sections \ref{section:Active_learning_digits} and \ref{sec:reject_curves}.
%
%\begin{equation}
%\label{eq:sumrule}
%   \textbf{q}(\textbf{x}) = \sum_i{\textbf{q}_{ij}(\textbf{x} | \textbf{x} \in C_{ij})}
%\end{equation}
%This should be followed by a normalization, such that the elements of $\textbf{q}(\textbf{x})$ (the individual class confidences) sum to one.

Several alternative updating schemes may be possible for a given a set of clusterings. In playing around with one of the datasets discussed in Section \ref{section:Hand-written digits} (the Block Letter dataset) the above procedure has been chosen for combining multi-level clusterings based on FMS or MS for active learning. We name it FMS-ALC, or MS-ALC if based on the MS clustering. The matrix A is large but sparse. In our implementation computations are made cluster by cluster avoiding the need to store A.
%
%\vspace{5mm}
%\begin{enumerate}
%  \item Select a set of neighborhood sizes and a complexity parameter.
%  \item Run the FMS algorithm with these parameters obtaining a set of clusterings $C$.
%  \item Rank them for increasing numbers of clusters.
%  \item Choose the clustering $C_0$ having a number of clusters for which all the modal objects can be labeled considering the cost of labeling.
%  \item Remove all clusterings with less clusters.
%  \item Initialize for all objects $q(\textbf{x}_t) = 0$ for all classes and $q(\textbf{x}_t) = 1$ for for the correct class of the labeled objects only.
%  \item Compute Equation \eqref{eq:updaterule} for all clusters and sequentially iterate over all clusterings using the obtained confidences of one clustering as initialization for the next one.
%  \item Assign every object $\textbf{x}_t$ to the class of the highest confidence in $\textbf{q}(\textbf{x}_t)$.
%\end{enumerate}
%\vspace{5mm}

\vspace{3mm}
\subsection{Reject curves for kNN mode seeking classification}
\label{section:reject}
As the active learning classifiers MS-ALC and FMS-ALC output confidences, they may be used for rejecting objects with a low confidence of the selected class (i.e. the class for which the corresponding element of $\textbf{q}(\textbf{x})$ is maximum). In Section \ref{section:Hand-written digits} some examples will be shown. In a sequential active learning scheme objects with a low classification confidence might be good candidates for additional labeling.

\vspace{3mm}
\subsection{Use of kMeans as a baseline procedure}
\label{section:kmeans}
A good candidate for a comparative study of the properties of the proposed FMS is the classical kMeans procedure \cite{Jain10}. It iteratively finds a preset number of clusters, $k$, by updating cluster means. In order to obtain modal objects, to label the clusters, the cluster procedure should be extended, e.g. by the computation of cluster medoid: objects in the clusters that are most close to the cluster means, see \cite{Kaufman90}. In our implementation medoid objects are just used for labeling, the clustering is not changed. The total number of distance computations is $(\eta k + 1) n$ if the algorithm runs for $\eta$ iterations. For large datasets, $\eta$ can be large before stability is reached. In practice the algorithm can be stopped prematurely obtaining an approximate clustering.

Note that the parameter $k$ has a different meaning in the two algorithms, kMeans and kNN mode seeking. In mode seeking $k$ determines the size of the neighborhood used for setting pointers to objects with a higher density. By following the pointers the total number of modes (clusters, objects to be labeled) in the entire dataset may be considerably smaller than $n/k$. The kMeans algorithm on the other hand will result in $k$ clusters (in some implementations, not in ours, a few clusters might be lost).

Like MS or FMS, kMeans can be used for active labeling. Instead of the modes the medoids are used. If the procedure is run for a number of times using different values of $k$, a similar multi-level clustering is found as by MS. Thereby two different classification procedures are obtained, kMeans-AL for active learning based on the medoids of a single clustering and kMeans-ALC if the set of clustering levels is combined.

Clusters found by the kMeans algorithm have a spherical shape. Moreover, all spheres have the same size. Consequently, the cluster medoids are, with some noise, uniformly distributed over the domain of the dataset. The MS procedure, on the other hand, finds clusters for which the shapes follow the shape of the density function around the density modes. In low density areas the clusters are thereby wide, in high density areas they are narrow.

A significant difference between the two approaches is the computational effort needed for finding a series of clusterings with different numbers of clusters. In MS this is performed in a single run, while kMeans needs for every cluster size a new call. Moreover, the time needed for such a call is proportional with $k$, the desired number of clusters. This implies that kMeans is bad for large numbers of clusters and good for small numbers. By comparing the numbers of distance computations needed for the two procedures, $2 n \sqrt{c n}$ for FMS (see Section \ref{section:fast algorithm}) and $(\eta k + 1) n$ for kMeans, see above, it shows that for  $k > \sqrt{n}$ kMeans has a larger computational complexity in case a single clustering with $k$ clusters has to be performed. In this reasoning we neglected the influence of the complexity parameter $c$ of FMS and the number of iterations $\eta$ used in of kMeans.

In case classification confidences are to be found, multiple clustering are needed, some with a high number of clusters. For such applications the time complexity of the kMeans procedure will be soon much larger than of FMS.

\subsection{Creating a nested set of multi-level clusterings}
\label{section:nesting}
For an arbitrary multi-level set of clusterings $C$ with $N_i$ clusters of clustering $C_i$ it is not guaranteed that the levels are nested. They are nested if all objects in a high-resolution cluster $C_{h j}$ belong to the same low-resolution cluster $C_{\ell k}$ with $N_h > N_{\ell}$. Nested clusterings can be better interpreted. E.g. agglomerative clusterings are automatically nested and can be inspected by a dendrogram.

There is an easy top-down procedure to make a given multi-level clustering nested if the clusters are represented by prototypes, like the modal objects in this paper. The following steps are used to change a given low-resolution clustering $C_{\ell}$ into a modified clustering $C'_{\ell}$ which is consistent with a high-resolution clustering $C_h$.

\vspace{5mm}
\begin{enumerate}
    \item All objects of all clusters in $C_h$ for which the cluster prototypes belong to the same cluster $C_{\ell j}$ are assigned to the same cluster in $C'_{\ell j}$.
    \item If the prototype of $C_{\ell j}$ is a member of $C'_{\ell j}$ it becomes the prototype of $C'_{\ell j}$ as well.
    \item If the prototype of $C_{\ell j}$ is not a member of $C'_{\ell j}$ the prototype of the largest constituting cluster in $C_h$ becomes the prototype of $C'_{\ell j}$.
\end{enumerate}
\vspace{5mm}

In our active learning experiments we extended kNN mode seeking and kMeans with nesting. The resulting procedures are called MS-ALN, FMS-ALN and kMeans-ALN.
%
%\subsection{Semi-supervised learning}
%\label{section:ssl}
%
%\emph{Skip or mention options for future research?
%}
%The proposed mode-seeking clustering procedures can be used for active learning, see Section \ref{section:active learning}, by assigning all unlabeled objects to the class of the modal objects found in a single clustering level. The improved classification procedure presented in Section \ref{section:multi-level_clustering} using all clustering levels, however, hardly depends on the fact that the given labeled objects are modal objects.
%
%If in step 4 of the algorithm the initial clustering level is chosen in relation with the size of a given labeled training set the algorithm is independent of usage of modal objects. It is just needed that most of the clusters contain one or more labeled objects. Consequently the presented classification procedure can be used generally to label a given unlabeled test set using a labeled training set by clustering the combined set and running the algorithm in \ref{section:multi-level_clustering}. We name these procedures FMS-SSL or MS-SSL. It should be realized that this results label estimated for the unlabeled objects used in clustering, including the class confidences, but not in a classifier that can be applied to new objects.
%
%In Section \ref{sec:ssl_experiments} some experiments are discussed.

\section{Application to hand-written characters}
\label{section:Hand-written digits}

In this section experiments will be presented with three related large datasets. They serve as an illustration of the possibilities of kNN mode seeking as well as an evaluation of its performance for clustering and active learning. They are selected on the basis of their size (70\,000 - 1.5 million objects) in order to show the speed of the algorithms for larger datasets.

\subsection{The datasets}
\label{section:datasets}

\begin{figure}[p]
\center
\includegraphics[height=5.5cm]{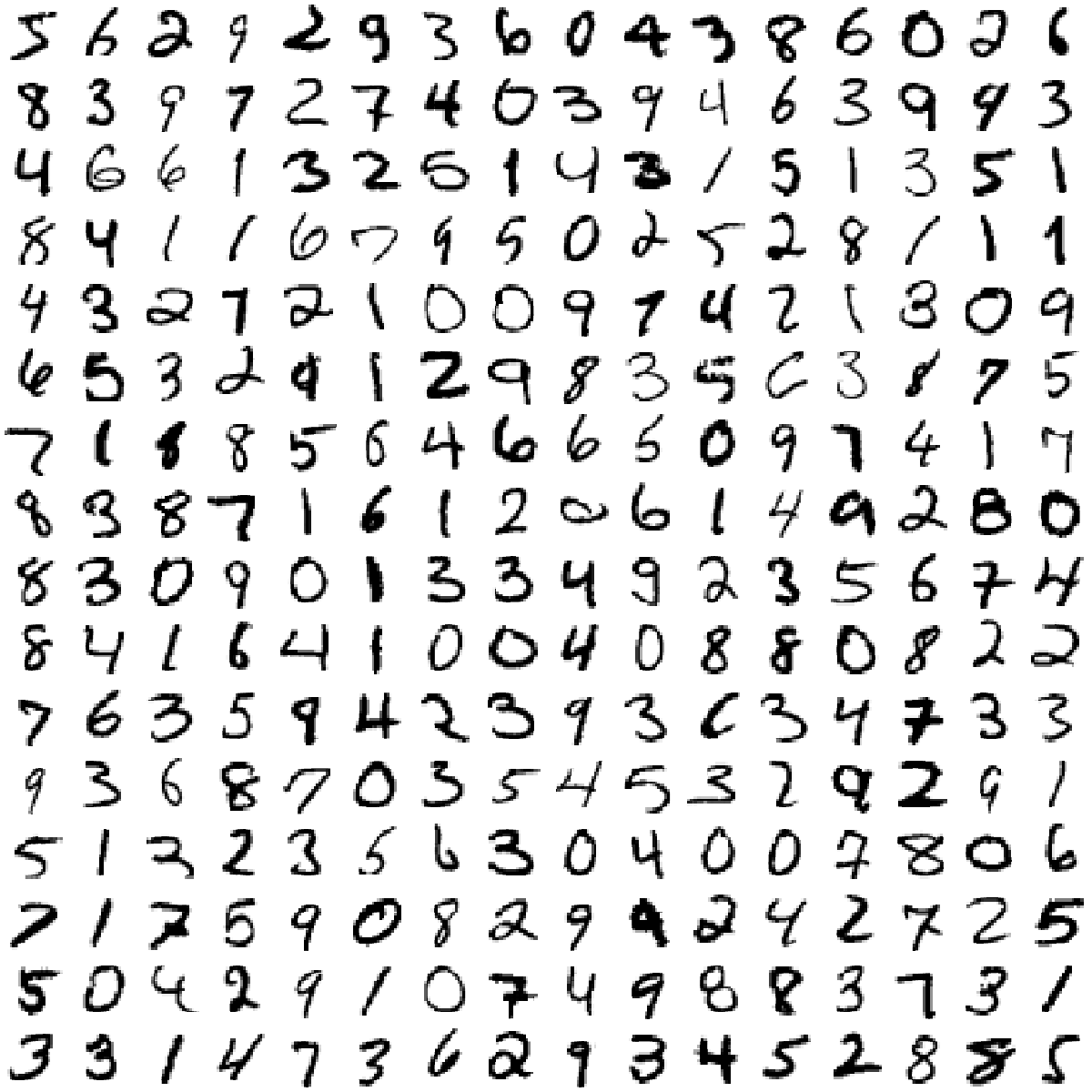} ~~
\includegraphics[height=5.5cm]{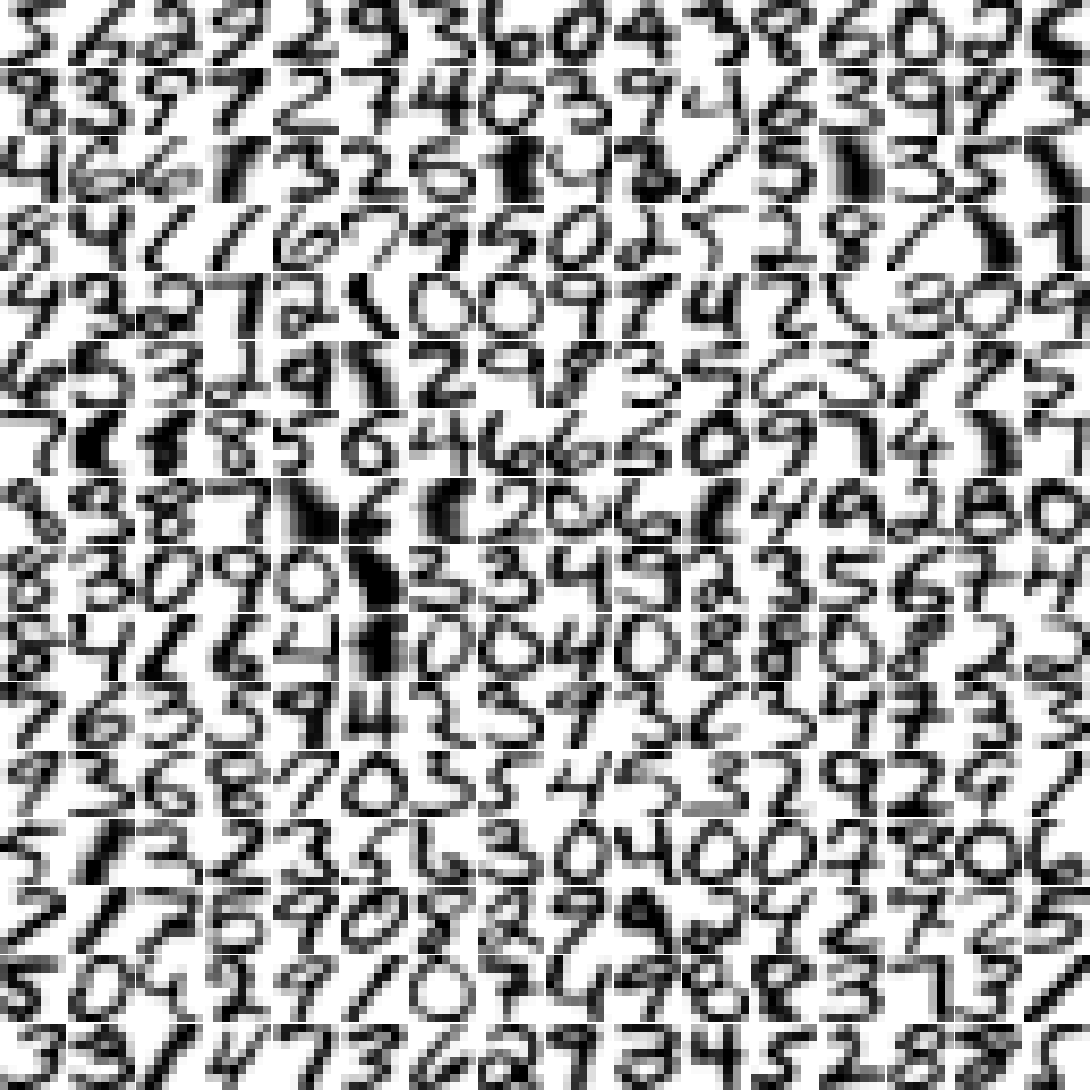}
\caption{MNIST examples, originals and normalized to $8 \times 8$ pixels.}
\vspace{-3mm}
\label{fig:CharsMNIST}
%\end{figure}

%\begin{figure}[!h]
\center
\includegraphics[height=5.5cm]{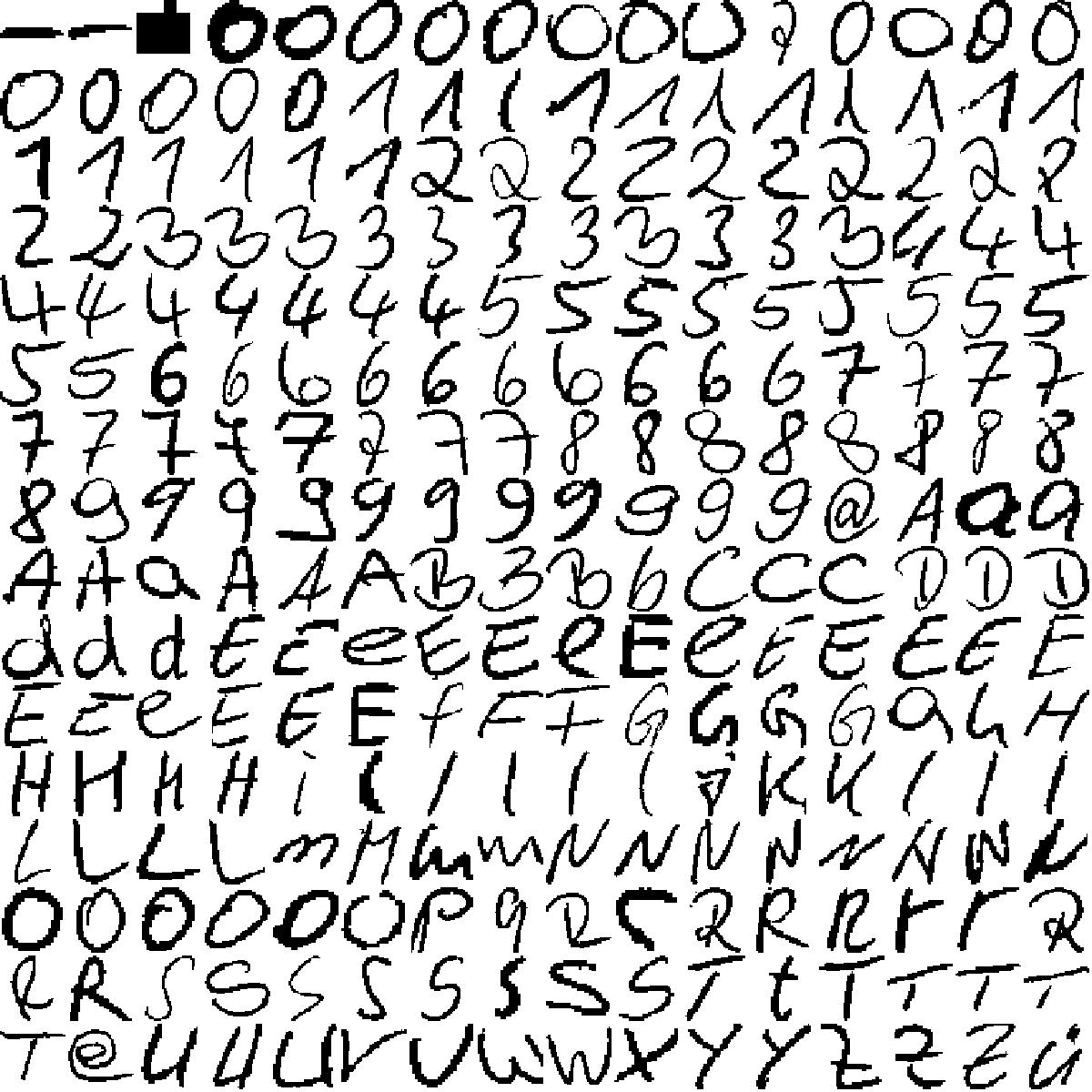} ~~
\includegraphics[height=5.5cm]{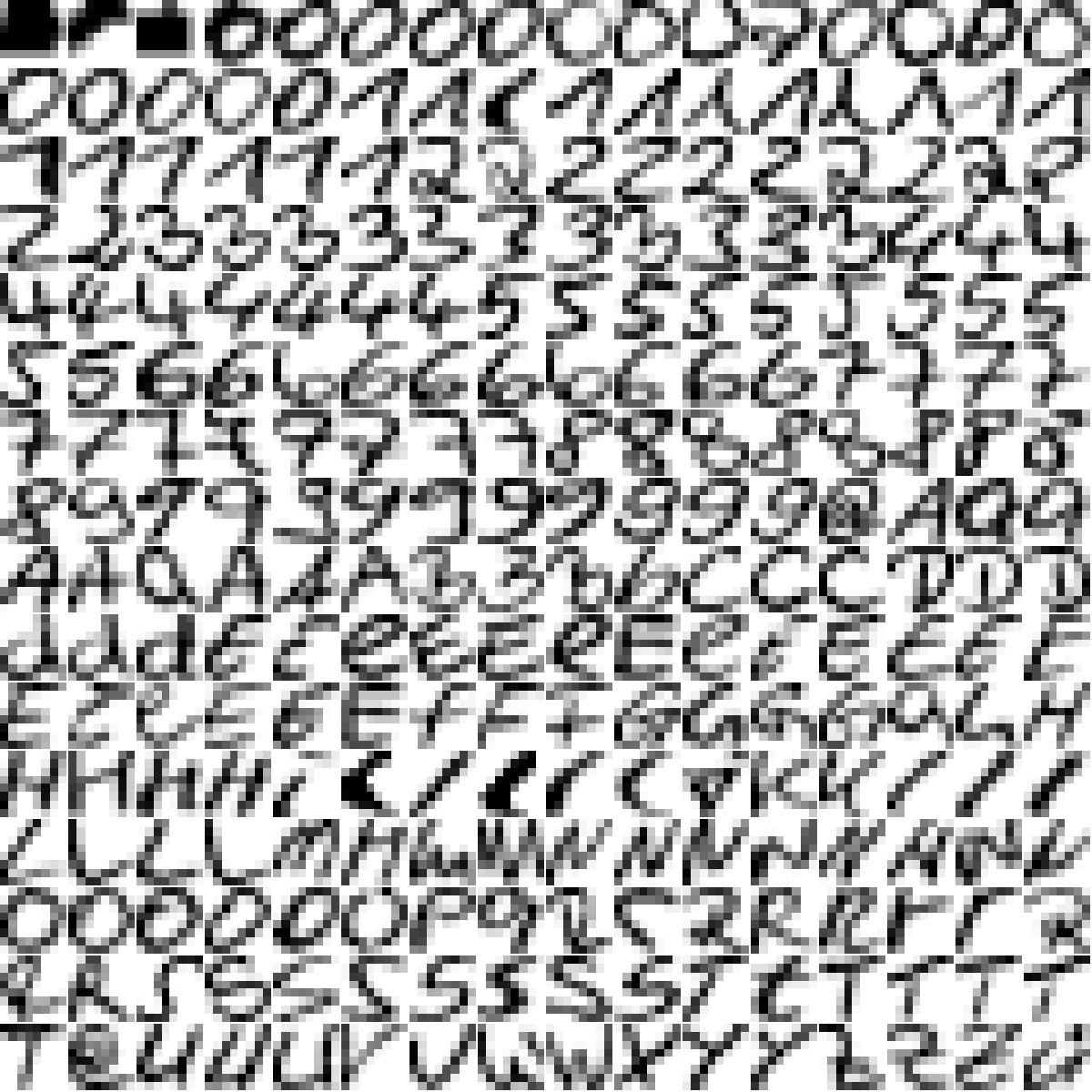}
\vspace{-3mm}
\caption{Block Letters examples, originals and normalized to $8 \times 8$ pixels.}
\label{fig:CharsBlock}
%\end{figure}

%\begin{figure}[!h]
\center
\includegraphics[height=5.5cm]{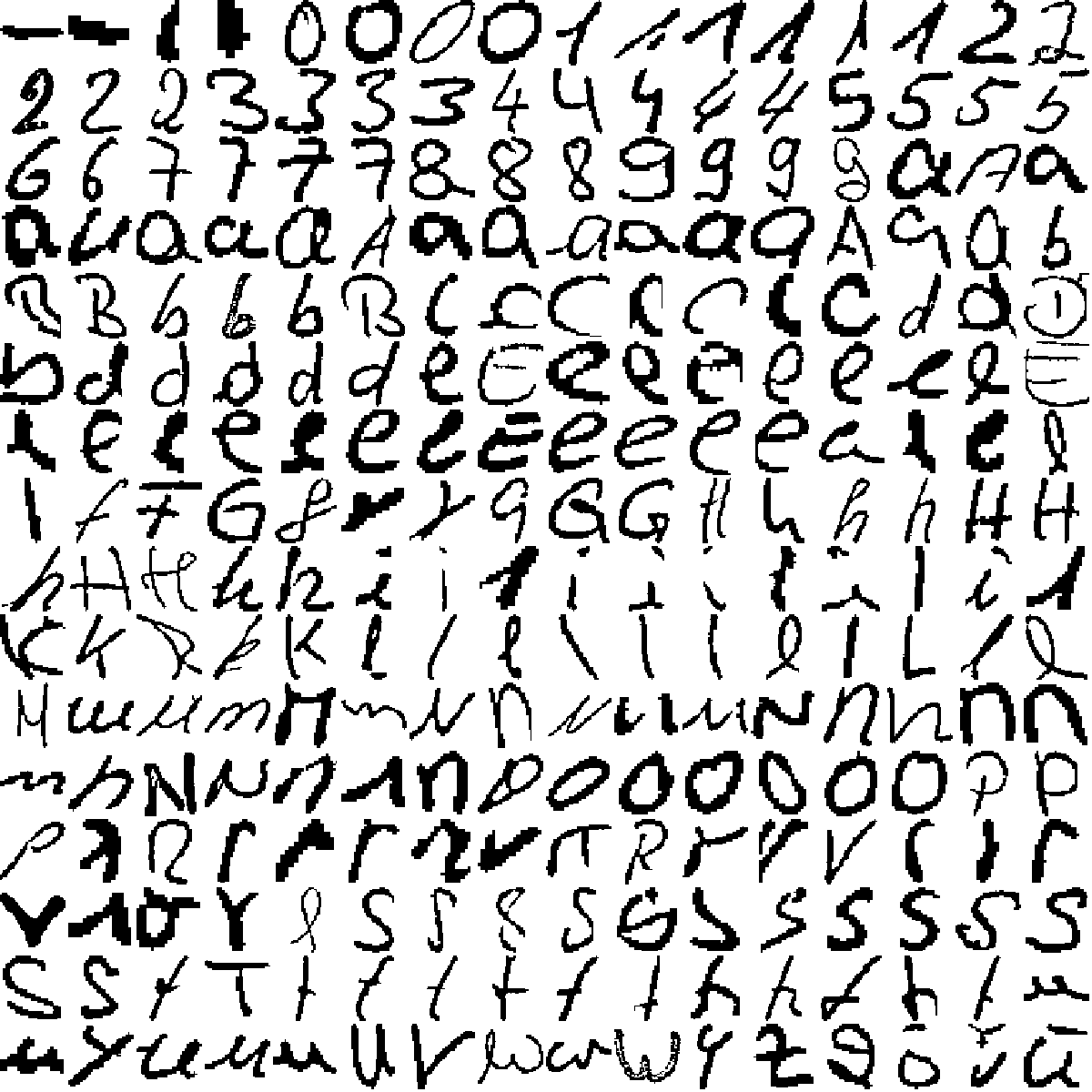} ~~
\includegraphics[height=5.5cm]{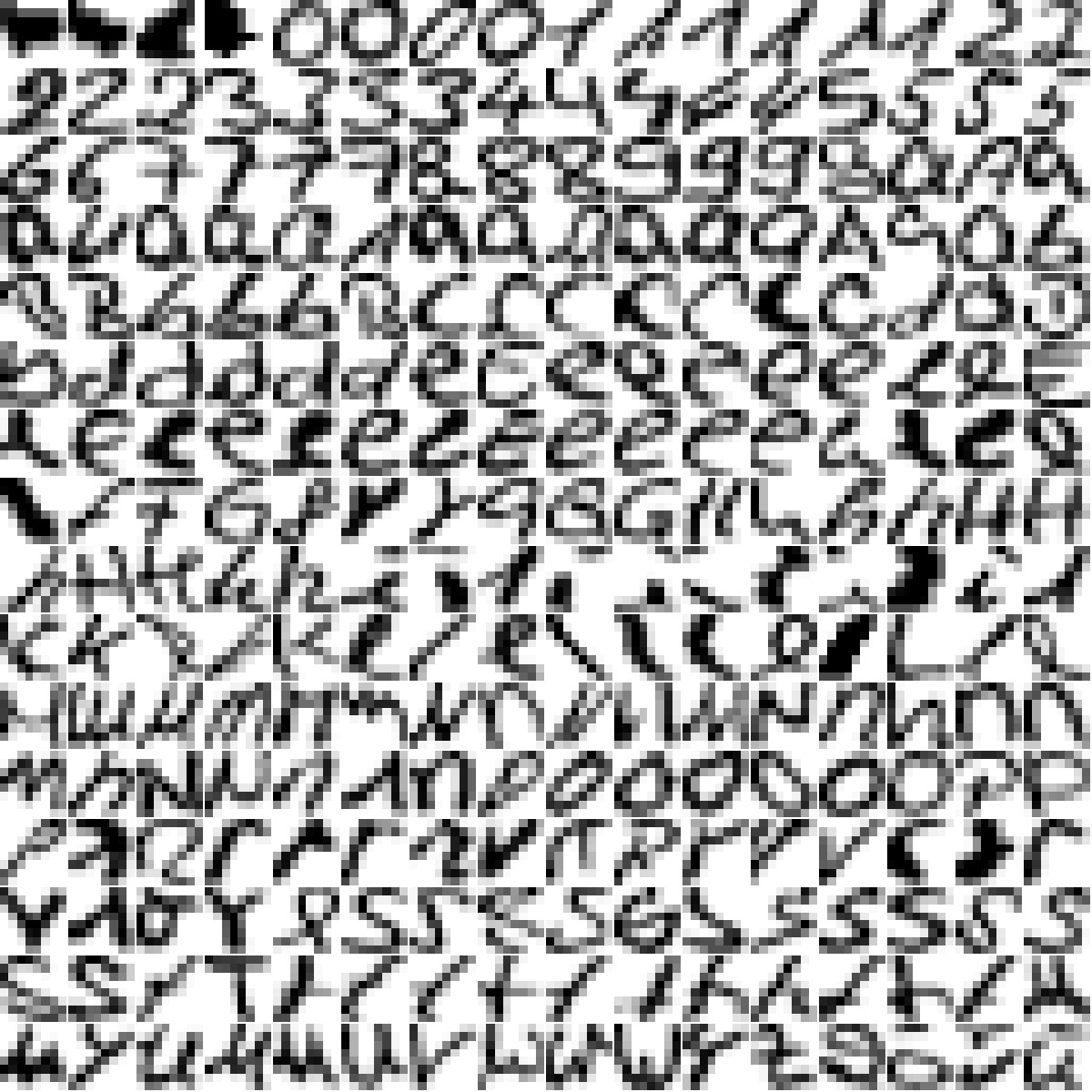}
\vspace{-3mm}
\caption{Cursive Letters examples, originals and normalized to $8 \times 8$ pixels.}
\label{fig:CharsCursive}

\end{figure}

We selected three large, real world character based datasets and represented them in the same way by their pixels, using the same classes. Additionally we combined the sets and enlarged the resulting set by including rotated copies of the characters. The tree basic sets are:
\begin{itemize}
  \item \textbf{MNIST}. This is the well-known public domain dataset of 70000 hand-written digits in 10 about equally sized classes. Originally they have the same size of 28 $\times$ 28 pixels. We normalized them all to 8 $\times$ 8 pixels using bilinear interpolation by which they can be represented in a 64-dimensional feature space. Feature vectors are normalized such that their elements sum to one. See Fig. \ref{fig:CharsMNIST} for an arbitrary subset, showing the originals as well as the normalized ones. For the purpose of this figure only, all sizes and maximum intensities are made equal. In a leave-one-out 1NN experiment a classification error of 0.025 was found for the original dataset and 0.020 for the normalized one.
  \item \textbf{Block Letters}. These are 82\,541 handwritten characters, labeled in 43 classes: letters, digits and some special symbols. The largest class contains 6\,109 characters, the smallest one has just 21 characters. The bounding boxes around the characters have different sizes, ranging from 3 $\times$ 3 to 65 $\times$ 70 pixels. We normalized this dataset in the same way as the MNIST digits, see Fig. \ref{fig:CharsBlock} for examples. In a leave-one-out 1NN experiment a classification error of 0.099 was found.
  \item \textbf{Cursive Letters}. These are 213\,623 handwritten characters, labeled in 42 classes, letters, digits and some special symbols. The largest class contains 21\,432 characters, the smallest one has 34 characters. The bounding boxes around the characters have different sizes, ranging from 2 $\times$ 2 to 375 $\times$ 318 pixels. We normalized this dataset in the same way as the MNIST digits, see Fig. \ref{fig:CharsCursive} for examples. In a leave-one-out 1NN experiment a classification error of 0.260 was found.
\end{itemize}

We also combined the three datasets into a single one, called \textbf{ALL}, with 366\,164 objects, having the same 43 classes as the Block Letters. In order to challenge the possibilities of the procedure even further this set is enlarged by a factor 4 by adding rotated versions of all characters over $\pi/2$, $\pi$ and $3\pi/2$ radials. This dataset, named \textbf{ALLR}, has 1\,464\,656 objects, still with 43 classes and represented by 64 features. The estimated leave-one-out 1NN errors of these two datasets are 0.230 and 0.237. The properties of the five datasets as used by us are summarized in Table \ref{tab:Datasets}.

\begin{table}
\center
\begin{tabular}{| l | c | r | c | c |}
\hline
Dataset & \#classes & \#objects & \#features & 1NN error\\
\hline
MNIST   & 10 &   70\,000 & 64 & 0.020 \\
Block   & 43 &   82\,541 & 64 & 0.099 \\
Cursive & 42 &  213\,623 & 64 & 0.260 \\
ALL     & 43 &  366\,164 & 64 & 0.230 \\
ALLR    & 43 & 1\,464\,656 & 64 & 0.237 \\
\hline
\end{tabular}
\vspace{3mm}
\caption{Summary of the datasets used in the experiments.}
\label{tab:Datasets}
\end{table}

\subsection{Clustering}
\label{section:clustering}

The following procedures, extensively discussed in the sections \ref{section:algorithm} and \ref{sec:classification} are used in the experiments:

\vspace{3mm}
\begin{itemize}
  \item \textbf{kMeans} The kMeans algorithm as described in \ref{section:kmeans}
  \item \textbf{MS} The original kNN mode seeking algorithm as described in \ref{section:old algorithm}
  \item \textbf{FMS} The fast kNN mode seeking algorithm as described in \ref{section:fast algorithm}. \textbf{FMS-2}, \textbf{FMS-4} and \textbf{FMS-6} stand for the variants with complexity parameters 2, 4 and 6.
\end{itemize}
\vspace{3mm}

\noindent
In the experiments with the MS procedures the set of neighborhood sizes ($K$) was chosen such that $K(i) = 1.21 \times K(i-1)$ with $K(1) = 2$ and $K(i) < n/10, \forall i$. $n$ is the number of objects in the dataset. Values in $K$ were rounded to integer. For $n = 100\,000$ the size of $K$ is 43, and for $n$ = 1.5 million (the size of ALLR) it is 57.

\begin{figure}[!ht]
%\begin{wrapfigure}{L}{0.55\textwidth}
\center
\includegraphics[width=0.65\textwidth]{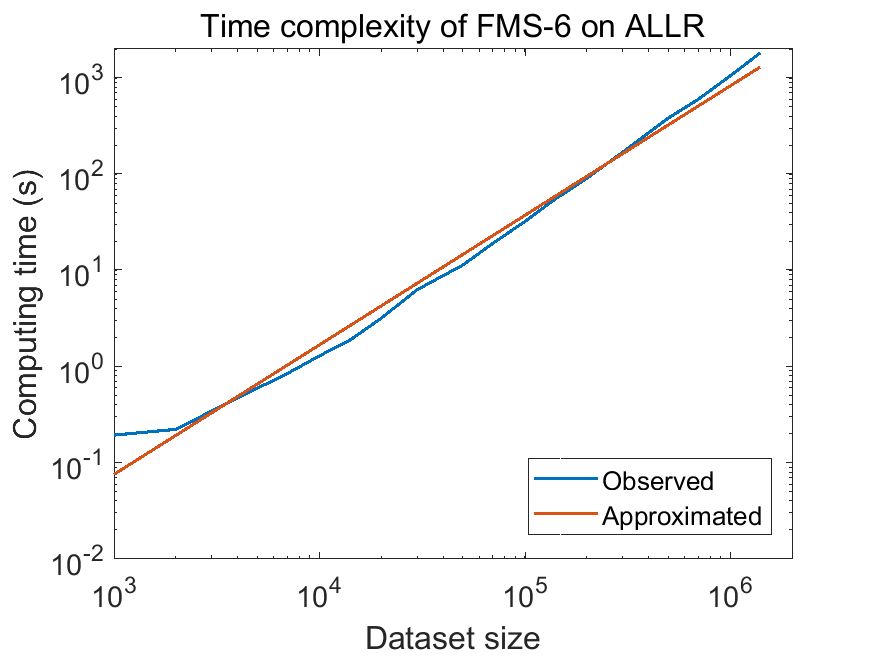}
\caption{Computing time for FMS-6 as a function of the dataset size. Subsets of the ALLR datasets were used. The figure shows that the time complexity is $\textsl{O}(n^{1.4})$.}
\vspace{-3mm}
\label{fig:TimeComplexity}
\end{figure}

First we used the largest dataset, ALLR, to verify the speedup that was expected for the FMS procedure. A complexity of $c = 6$ was used in the construction of the Q-cells, see Section \ref{section:fast algorithm}. Computing times for various subsets of the ALLR dataset are shown in Figure \ref{fig:TimeComplexity}. Surprisingly it shows that the computing time is $\textsl{O}(n^{1.4})$ while it was argued in Section \ref{section:fast algorithm} that the minimum would be $\textsl{O}(n^{1.5})$. Our explanation is that smaller datasets have smaller P-cells and Q-cells, resulting in smaller matrices used in the distance computation. In our Matlab implementation this is less efficient. The clustering of larger datasets can thereby be computed more efficiently.

In order to study the influence of the complexity parameter of the FMS algorithm (see Section \ref{section:fast algorithm}) it was run on the three smaller datasets, MNIST, Block and Cursive for $c$ = 2, 4 and 6. Results are compared with MS and kMeans. The latter was run for all numbers of clusters that were found by MS. To save computing time of kMeans, at the cost of accuracy, optimization iterations have been limited to 10s for every number of clusters. Computing times in seconds on a 2.66 GHz HP Z4000 PC under Windows-7 with 8 Gbyte memory are given in table \ref{tab:Datasets clustered}.

\vspace{5mm}
\begin{table}[!h]
\begin{tabular}{| l | r | r | r | r | r | r |}
\hline
Dataset & \#objects & kMeans & MS & FMS-2 & FMS-4 & FMS-6\\
\hline
MNIST   &   70000 &  314 &  1000 &  8.9 &  11.9 &  14.6 \\
Block   &   82541 &  398 &  1460 &  9.4 &  13.1 &  16.6 \\
Cursive &  213623 &  634 & 10300 & 44.7 &  63.8 &  83.0 \\
ALL     &  366164 & 1420 & 30580 & 97.4 & 147   & 187 \\
ALLR    & 1464656 &  &   &   870 & 1252 & 1590\\
\hline
\end{tabular}
\vspace{3mm}
\caption{Computing times in seconds for a set of multi-level clusterings, see also Figure \ref{fig:Clustering_Evalution}.}
\label{tab:Datasets clustered}
\end{table}

\begin{figure}[!p]
\center
\includegraphics[width=6.5cm]{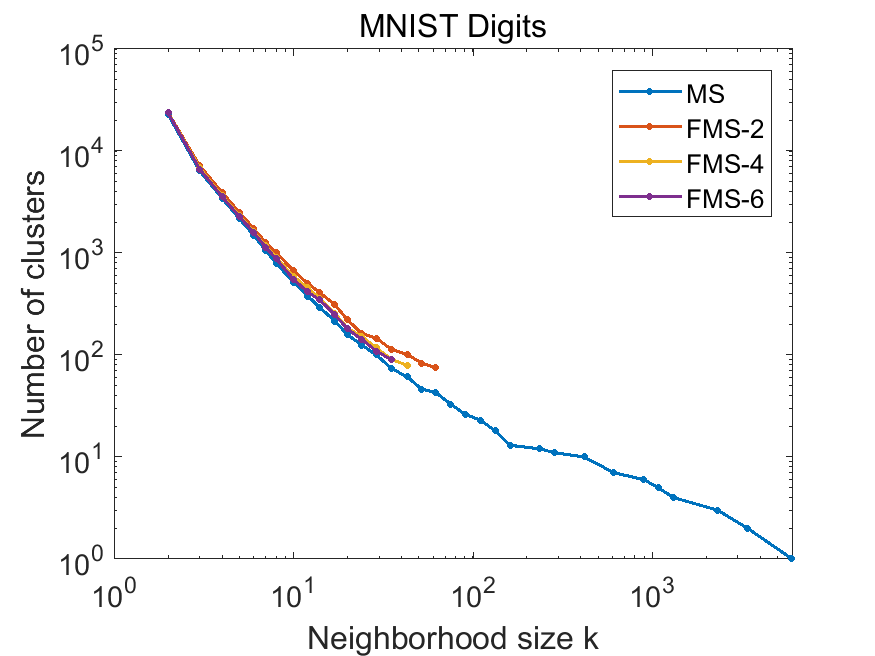} ~~
\includegraphics[width=6.5cm]{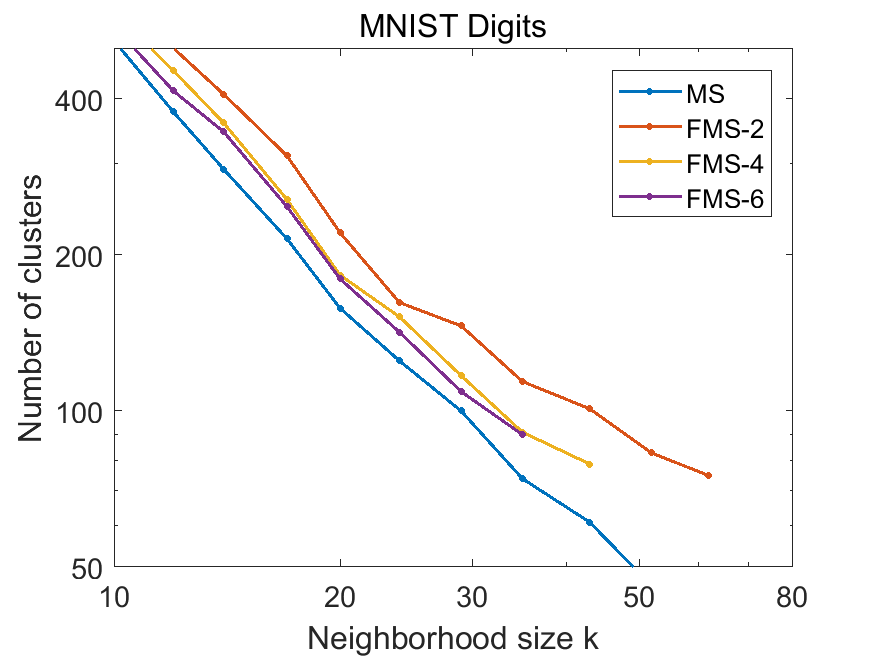}\\
\vspace{5mm}
\includegraphics[width=6.5cm]{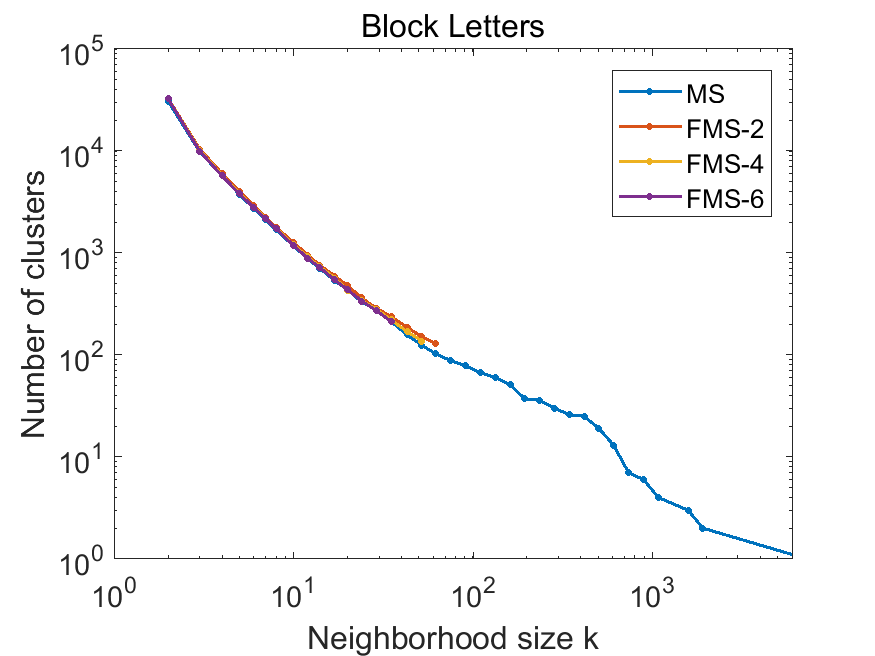} ~~
\includegraphics[width=6.5cm]{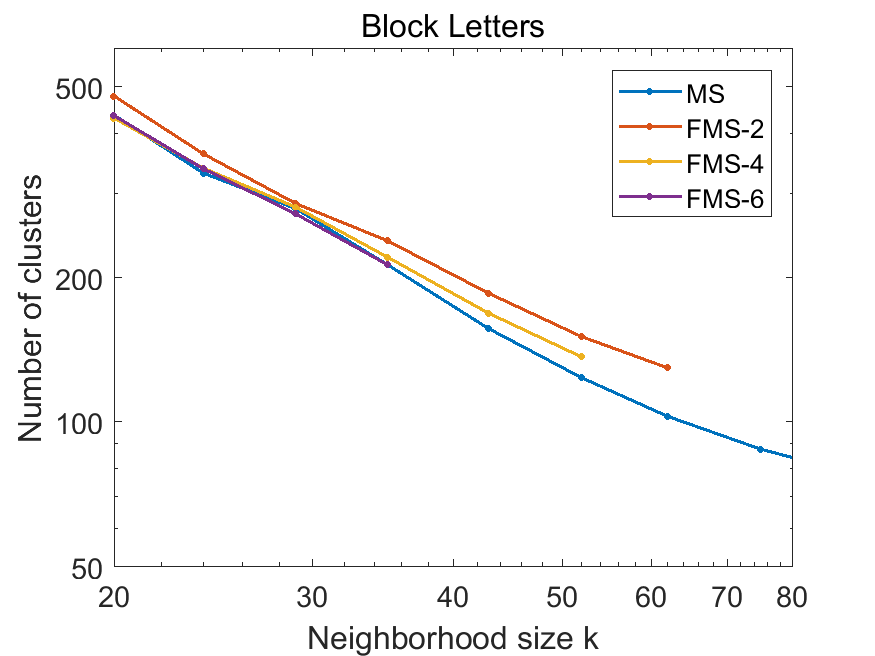}\\
\vspace{5mm}
\includegraphics[width=6.5cm]{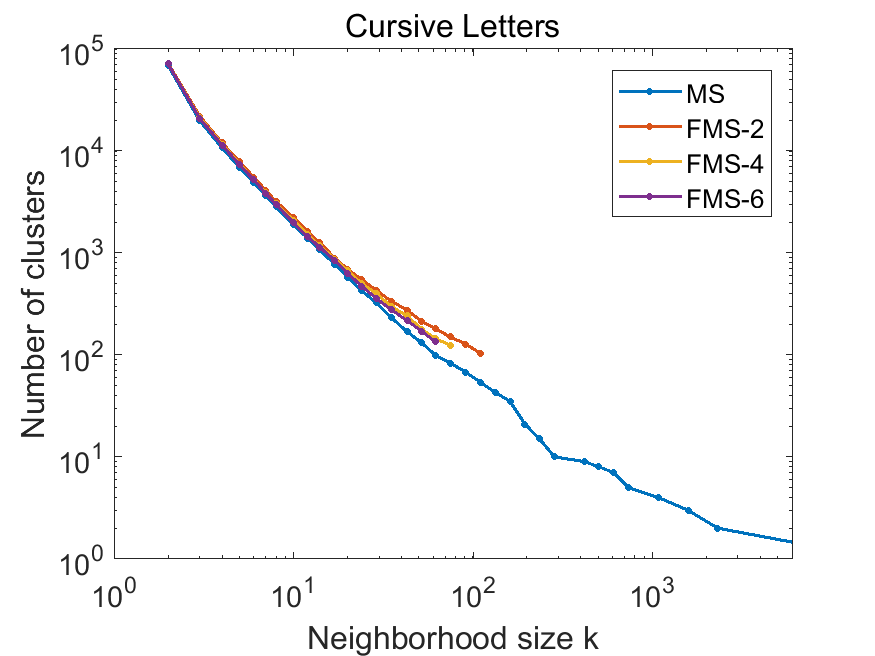} ~~
\includegraphics[width=6.5cm]{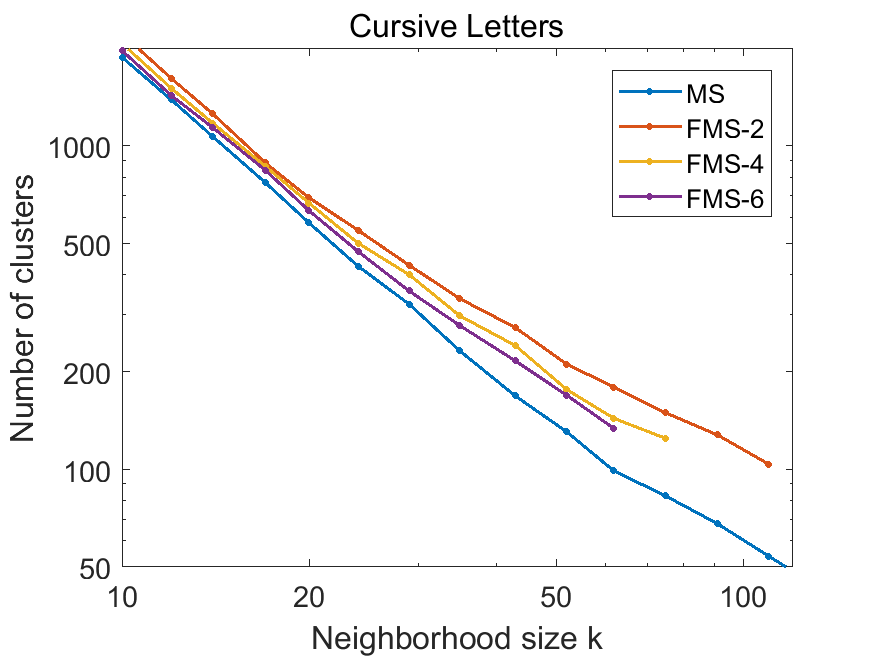}\\
\caption{Numbers of clusters found for the three datasets by MS and FMS. The right figures are zoomed versions of the ones on the left.}
\vspace{-3mm}
\label{fig:Cluster_numbers}
\end{figure}

In Figure \ref{fig:Cluster_numbers} the numbers of clusters found by MS and FMS are shown as a function of the neighborhood parameter $k$ for the three smaller datasets. Two observations may be of interest. First, the figures show that the fast procedures FMS-2, FMS-4 and FMS-6 find for small neighborhood sizes about the same numbers of clusters as found by the slow original procedure MS. For $k \approx 20$ the numbers slowly deviate: more clusters are found than by MS most likely due to the limited cell sizes by which some clusters cannot be merged. It can be clearly observed that the higher the complexity parameter $c$ the better the FMS-$c$ curves approximate the MS ones. For $k \approx 70$ clustering by FMS stops as this is about the size of the smallest cells.

%\begin{figure}[!h]
\begin{wrapfigure}{L}{0.55\textwidth}
\center
\includegraphics[width=0.55\textwidth]{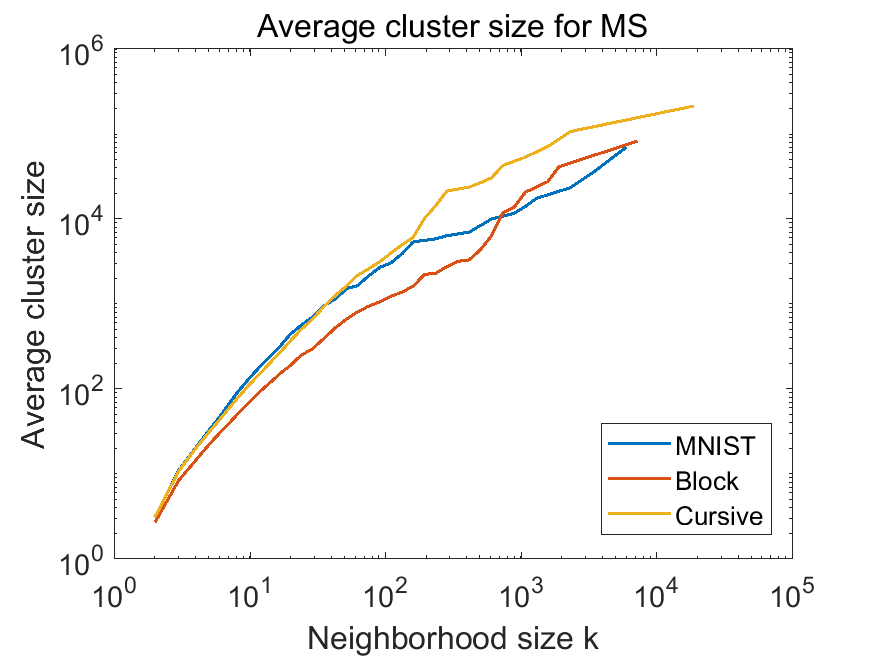}
\caption{Averaged cluster sizes of the MS procedure for the three smaller datasets.}
\label{fig:Clustersizes}
\end{wrapfigure}

A second observation is that the average cluster sizes of the three datasets seem to be different. In a combined graph based on the MS curves, Figure \ref{fig:Clustersizes}, this can be observed better. The Block Letter dataset has smaller clusters than the Cursive Letters as it has considerably less objects and is thereby less dense. The even smaller MNIST Digits dataset has, however, for $k < 200$ about the same cluster sizes as Cursive Letters. This might be explained by the lower variability of MNIST (just ten well defined classes), see Figure \ref{fig:CharsMNIST}.

\begin{figure}[!p]
\center
\includegraphics[width=6.5cm]{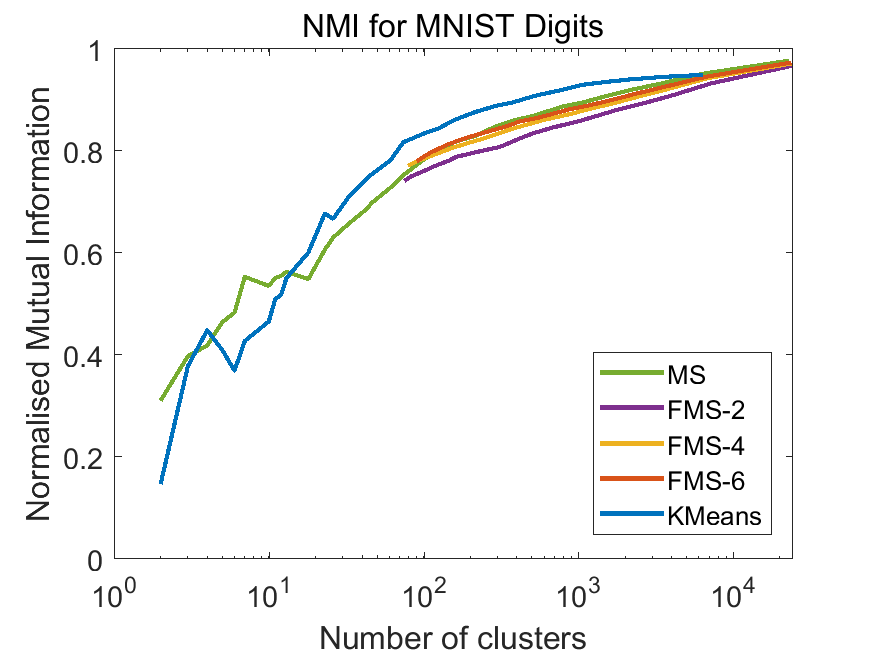} ~~
\includegraphics[width=6.5cm]{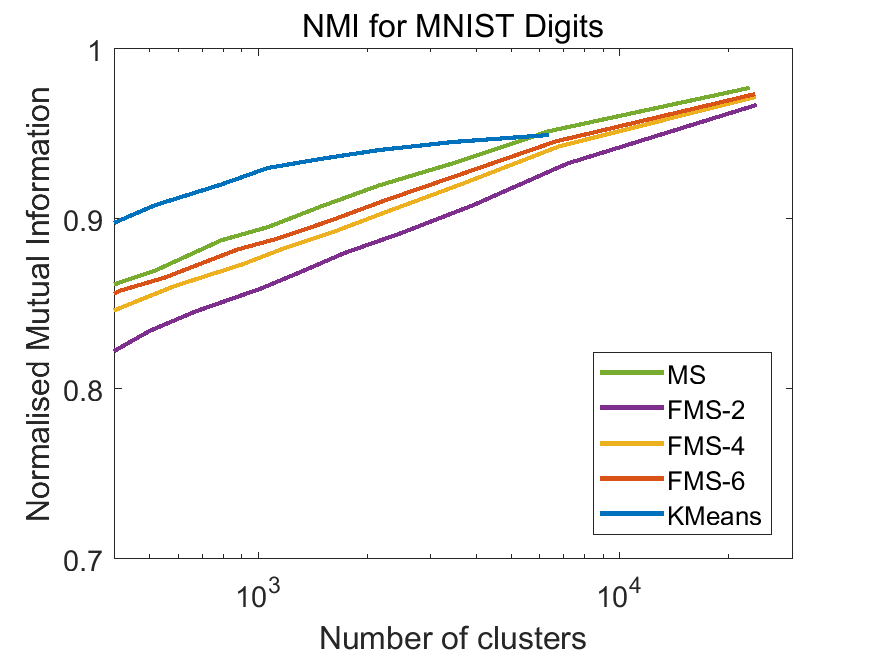}\\
\vspace{5mm}
\includegraphics[width=6.5cm]{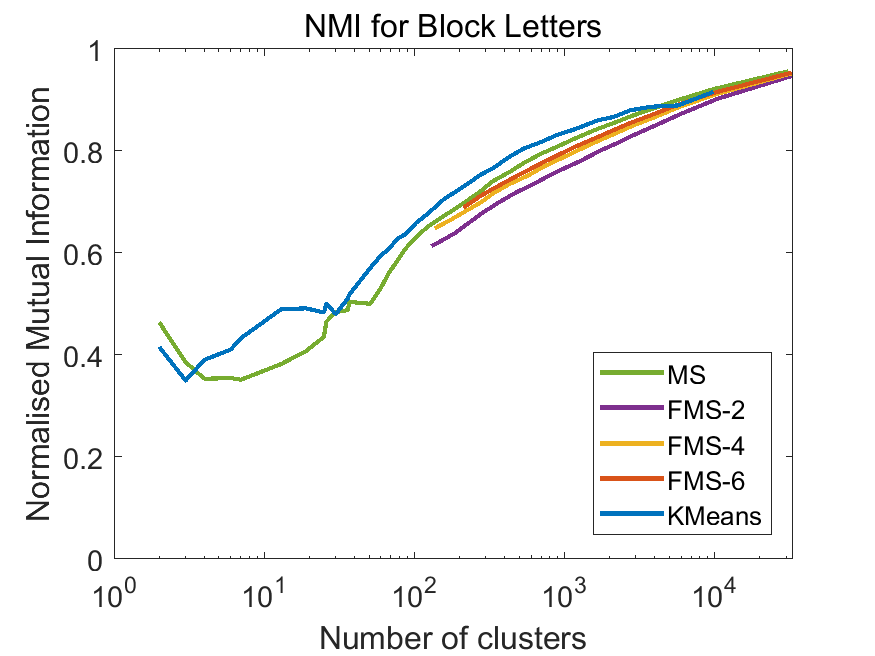} ~~
\includegraphics[width=6.5cm]{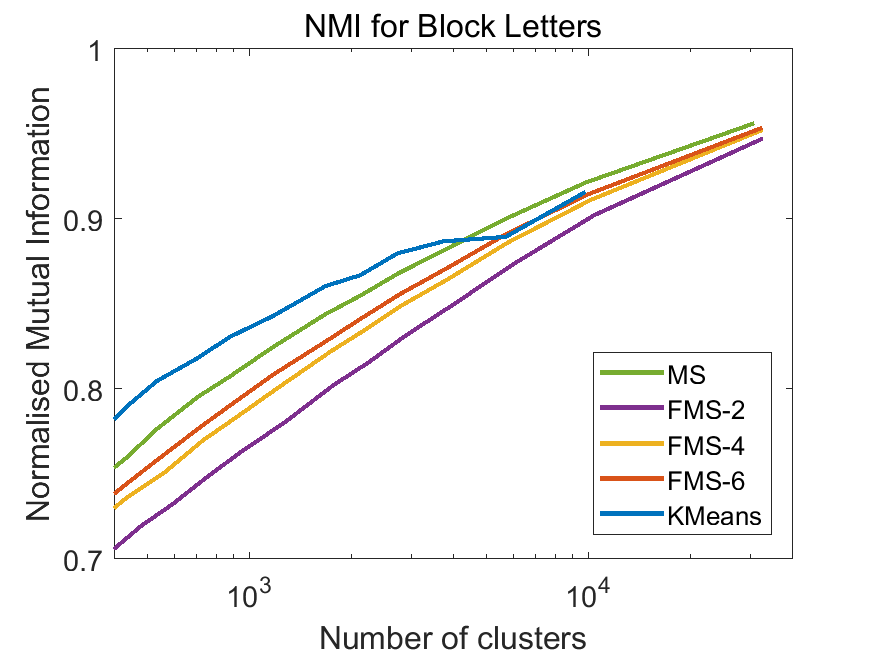}\\
\vspace{5mm}
\includegraphics[width=6.5cm]{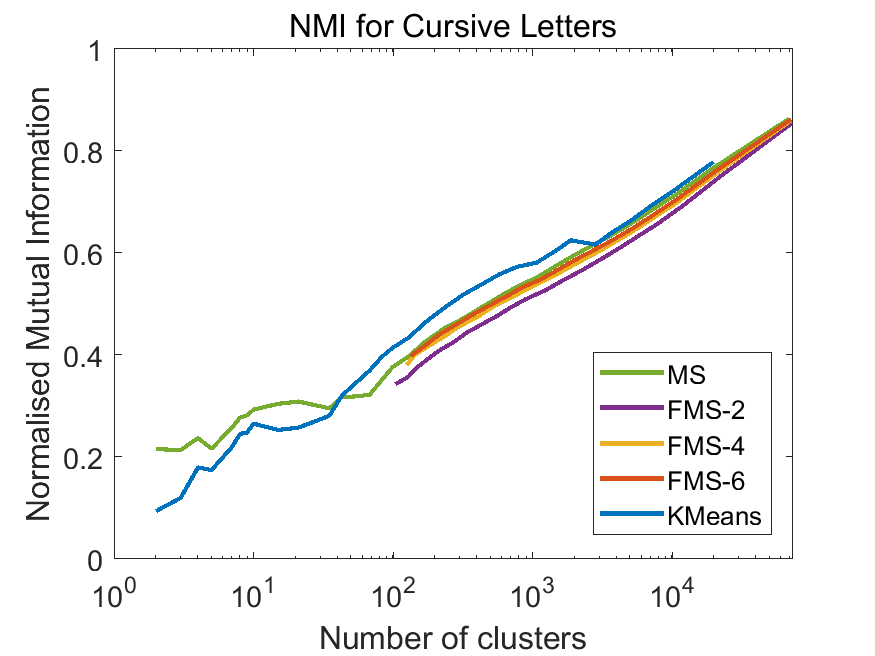} ~~
\includegraphics[width=6.5cm]{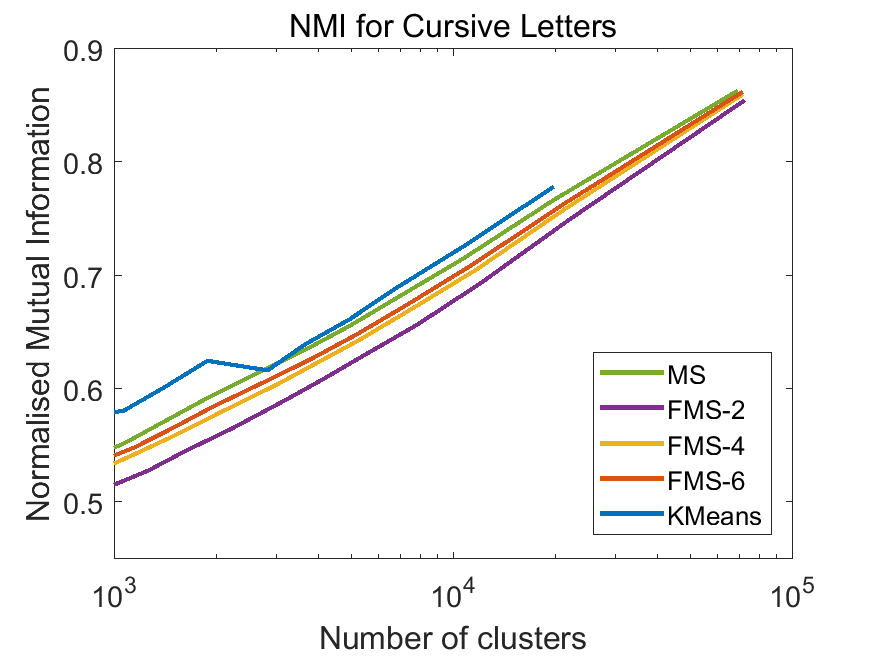}\\
\caption{Clustering evaluation for three datasets based on the Normalized Mutual Information (NMI) as defined by \eqref{eq:NMI}. The right figures are zoomed versions of the ones on the left.}
\vspace{-3mm}
\label{fig:Clustering_Evalution}
\end{figure}

\subsection{Cluster evaluation by the mutual information with true labels}
\label{section:cluster_evaluation}

The purpose of clustering is to find meaningful subsets in the data. Whether clusters are meaningful is application dependent, but very often it implies that they should make sense for human judgement. In this paper we are thereby using labeled datasets and verify to what extend the obtained clusters correspond with the human class labeling. The measure we use is the below defined Normalized Mutual Information (NMI) between the cluster labels $\eta_i$ and the class labels $\lambda_j$ (note that the number of classes is usually different from the number of clusters)

\begin{equation}
\label{eq:MI}
I(\eta,\lambda) = \sum_{\forall i} \sum_{\forall j} \left\{ p(\eta_i,\lambda_j)\log{ \left( \frac{p(\eta_i,\lambda_j)}{p(\eta_i)p(\lambda_j)} \right) } \right\}
\end{equation}
which is the mutual information. It is normalized to obtain values between 0 and 1:

\begin{equation}
\label{eq:NMI}
I_n(\eta,\lambda) = \frac{I(\eta,\lambda)}{\min\{H(\eta),H(\lambda\}}
\end{equation}

%\begin{figure}[!h]
\begin{wrapfigure}{L}{0.55\textwidth}
\center
\includegraphics[width=0.55\textwidth]{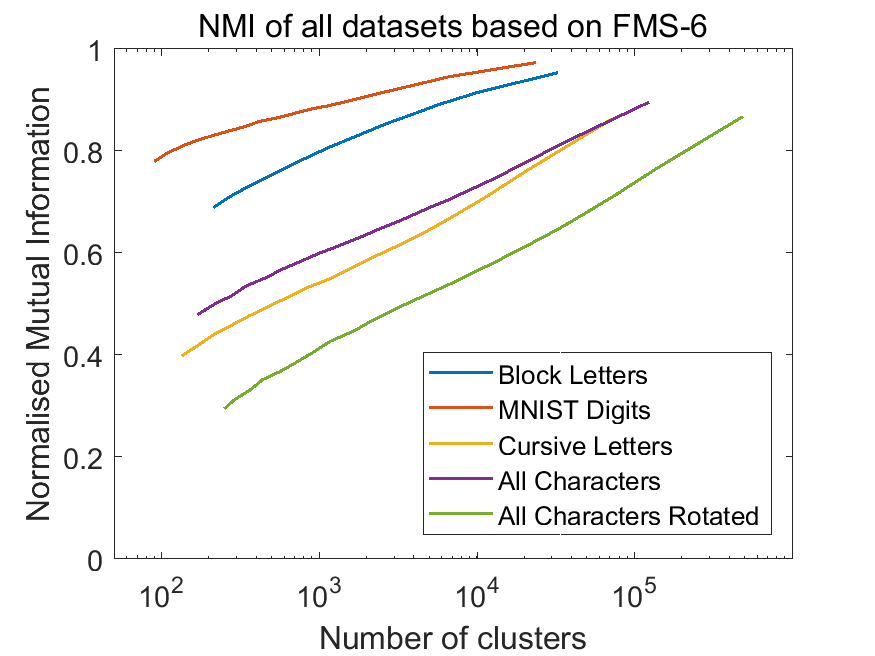}\\
\caption{Clustering evaluation by the Normalized Mutual Information (NMI) for all datasets based on FMS-6.}
\label{fig:Clustering_Evalution All}
\end{wrapfigure}

\noindent
in which $H(x) = -\sum_{\forall x}\{p(x)\log{p(x)}\}$ is the entropy of a discrete random variable $x$. The values of NMI \eqref{eq:NMI} between the true class labels and the MS and FMS clusterings of three datasets is shown in Figure \ref{fig:Clustering_Evalution}. For large numbers of clusters the value of NMI approximates one as small clusters tend to have objects of a single class. kMeans performs similar and often even better than MS on the basis of this criterion.  For larger complexities $c$ results for the proposed FMS-$c$ are almost similar to those of the much slower original MS algorithm.

In Figure \ref{fig:Clustering_Evalution All} the evaluation curves of the five datasets are shown together. They all grow asymptotically to NMI = 1, which is obviously the case if the number of clusters equals the size of the dataset. If every object is in a separate cluster the mutual information between the object labels and the clustering is one. The MNIST results are better as there are just 10 instead of 43 classes and because these classes are better separable. Moreover, in general holds that the larger the dataset the lower the curves. See Table \ref{tab:Datasets} for dataset sizes.

\subsection{Cluster evaluation by active learning}
\label{section:Active_learning_digits}

\begin{figure}[!p]
\center
\includegraphics[width=6.5cm]{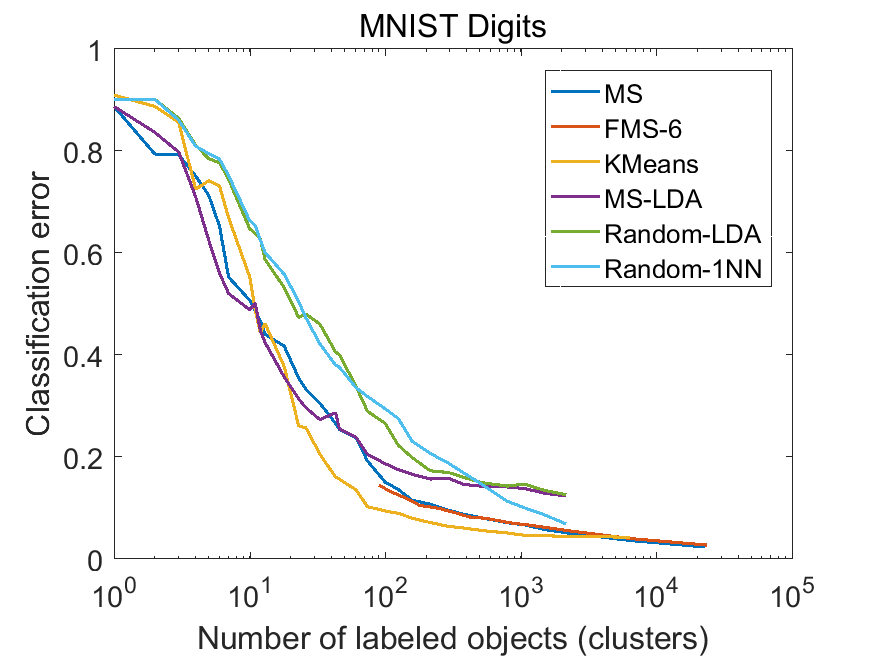} ~~
\includegraphics[width=6.5cm]{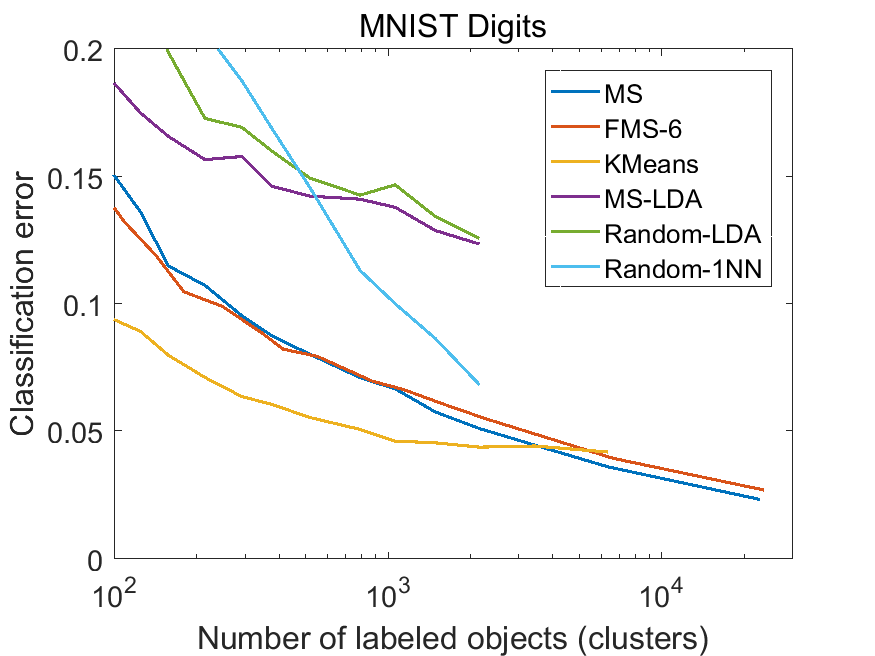}\\
\vspace{5mm}
\includegraphics[width=6.5cm]{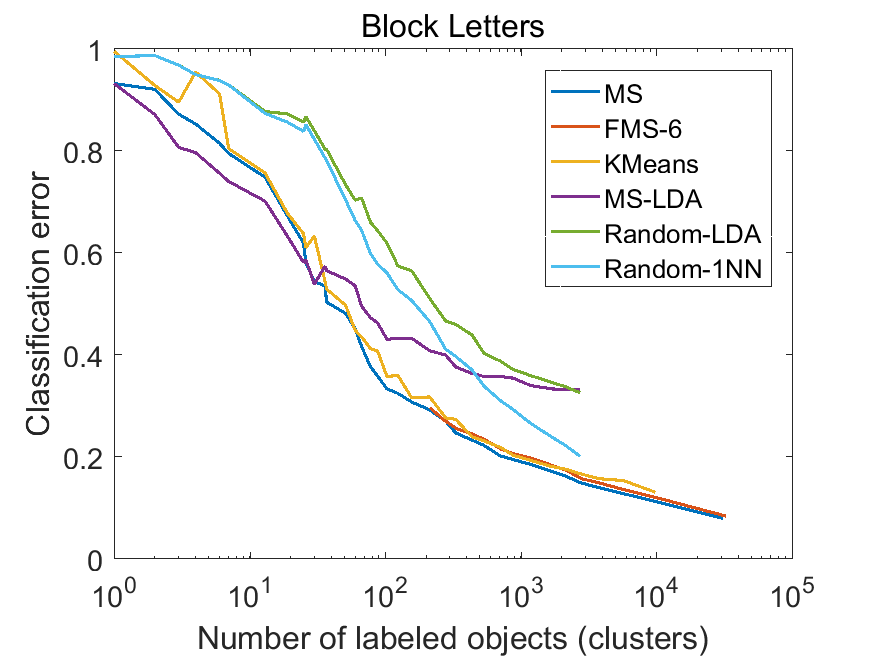} ~~
\includegraphics[width=6.5cm]{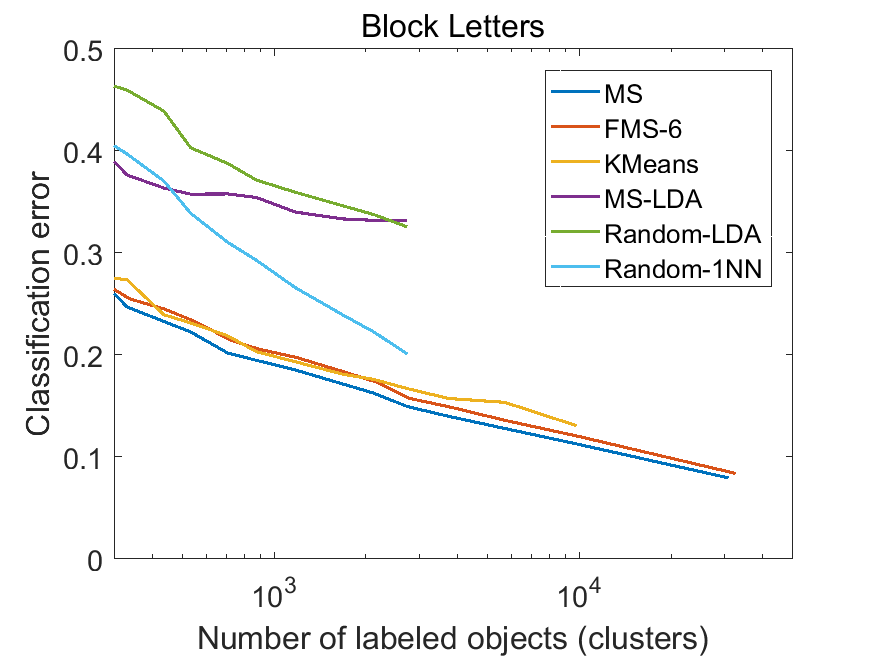}\\
\vspace{5mm}
\includegraphics[width=6.5cm]{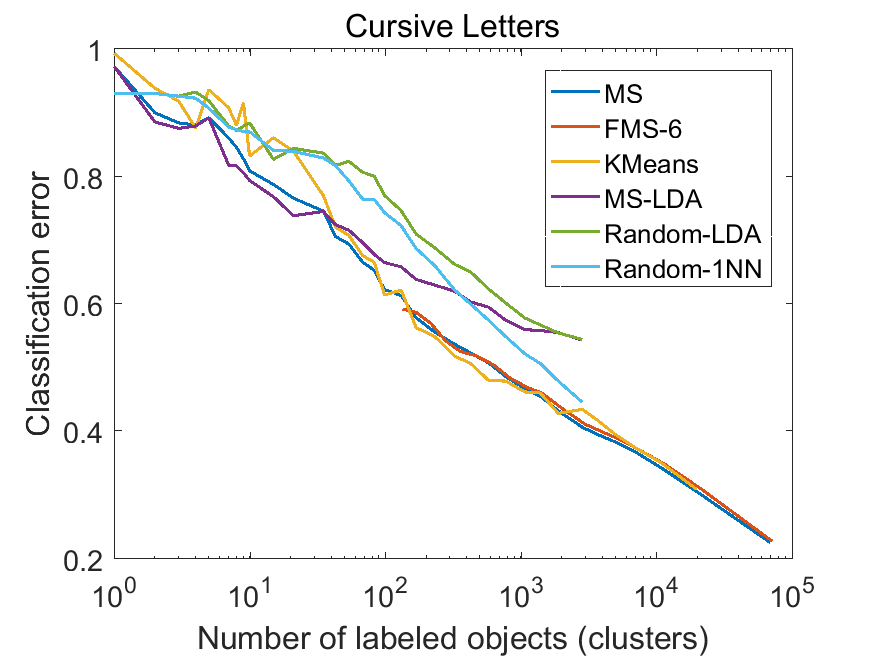} ~~
\includegraphics[width=6.5cm]{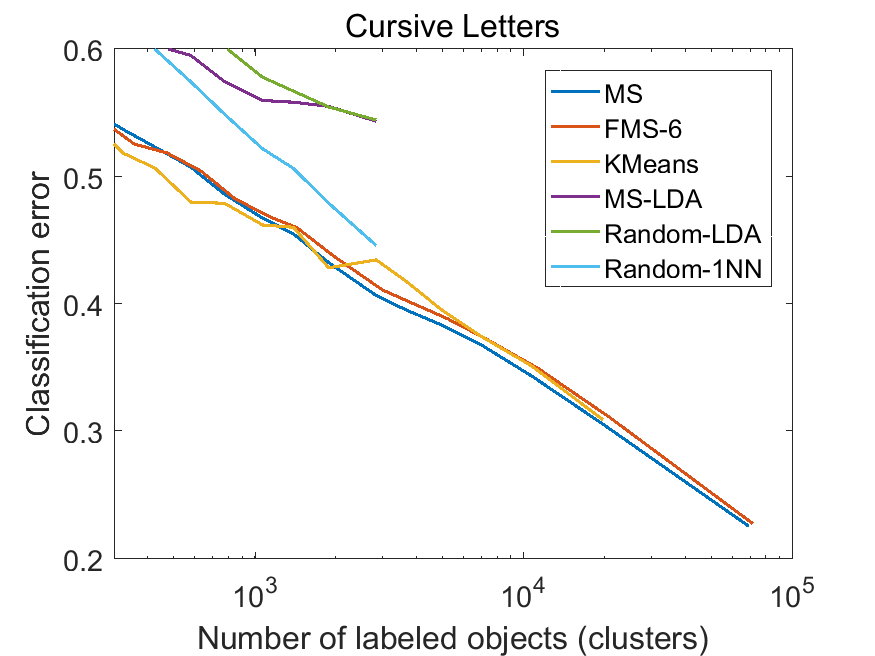}\\
\caption{Active Learning evaluation for three datasets. MS-LDA is the LDA classifier trained by the modal objects of the clusters. Random-LDA and Random-1NN are trained by a randomly selected training set with the size of the number of clusters on the x-axis. The right figures are zoomed versions of the ones on the left.}
\vspace{-3mm}
\label{fig:ActLearning_Evaluation}
\end{figure}

An important application of clustering of large datasets is active learning: representative objects to be labeled are found by the structure in the data. In the next step these objects may be used for training a classifier. Alternatively the clusters can be used to label the unlabeled objects by the labeled objects in the same cluster. In our experiments the modal objects obtained by MS or FMS or the medoid objects determined in the kMeans clusters are used for labeling the other objects in the same cluster.

In Figure \ref{fig:ActLearning_Evaluation} learning curves are presented for the three clustering procedures, kMeans, MS and FMS-6. They are compared with two LDA classifiers and the 1NN rule:
\begin{itemize}
  \item MS-LDA, using the modal objects of the clusters, after retrieving their labels, for training.
  \item Random-LDA, using a randomly generated training set, repeated 10 times and averaged.
  \item Random-1NN, using a randomly generated training set, repeated 10 times and averaged.
\end{itemize}

These curves are based on single experiments and look noisy. A number of them, those based on MS, are deterministic as they are not dependent on a randomly selected subset of objects. So repeating and averaging will not help. Nevertheless, it is clear that the clustering based classifications are significantly better than those obtained by the LDA and 1NN. This can be expected as these classifiers depend on the selected objects only, while the clustering based classifications use the entire data structure. An advantage of training a classifier, however, is that it can be used for future objects while the clustering based classification is only applicable to the objects that participated in the clustering.

There is a distinction between the two LDA classifiers, suggesting that using a clustering scheme for selecting a training set for a classifier is better than using a randomly selected training set.

\begin{wrapfigure}{L}{0.50\textwidth}
\center
\includegraphics[width=0.50\textwidth]{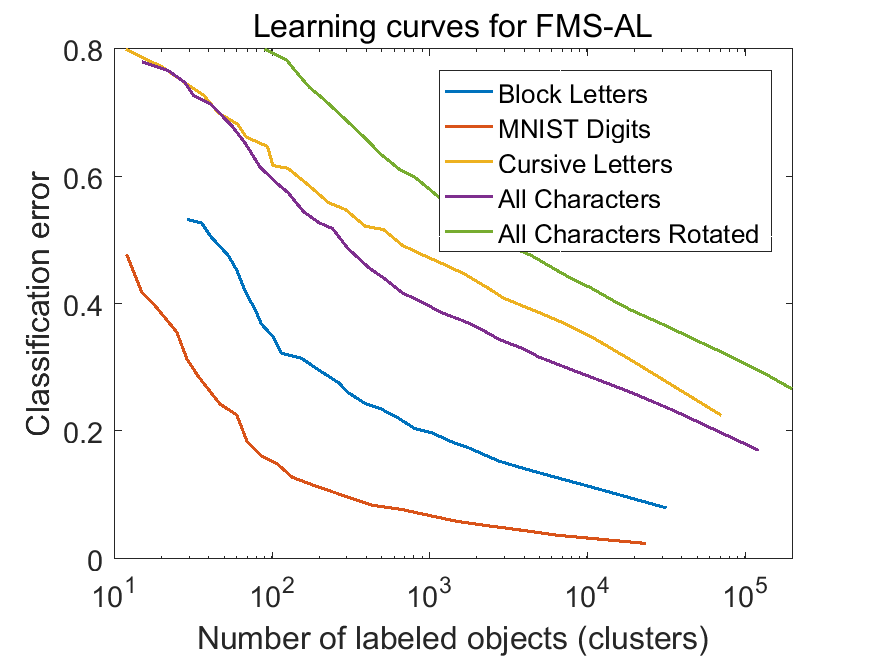}
\caption{Active Learning evaluation based on FMS-AL compared for all datasets.}
\label{fig:ActLearning_Evaluation Fast}
\end{wrapfigure}

As was observed for the mutual information in Section \ref{section:cluster_evaluation}, it appears that also for active learning the fast algorithm FMS-$c$ approximates the performance of the original procedure MS for larger complexities, see Figure \ref{fig:ActLearning_Evaluation}. The results of the maximum complexity $c = 6$ studied here are close to the results of MS, but their computation is some orders of magnitude faster, see Section \ref{section:clustering} and Table \ref{tab:Datasets clustered}. In Figure \ref{fig:ActLearning_Evaluation Fast} the learning curves of FMS-6 based on the classification error are summarized for all datasets. The relative behavior is similar to the mutual information curves in Figure \ref{fig:Clustering_Evalution All}.

The kMeans procedure is sometimes better, sometimes worse than the MS based procedures. However, the computation of learning curves, as presented here, is highly computational intensive for kMeans, as it has to be repeated for all values of $k$. The mode seeking clustering MS needs just a single run and can be well approximated by the fast FMS algorithm, see Table \ref{tab:Datasets clustered}. Recall that the number of distance computations needed for FMS is $\textsl{O}(n \sqrt(n))$, while it is for kMeans $\textsl{O}(n k \eta)$ with $\eta$ the number of iterations used per update. As we have bounded the computing time per iteration to 10s it appeared that for small values of $nk$ hundreds of iterations could be used, while for large values (for our data about $nk > 2 \times 10^8$) just one update could be possible. This may have deteriorated the results of kMeans.

\begin{figure}[!h]
\center
\includegraphics[width=6.5cm]{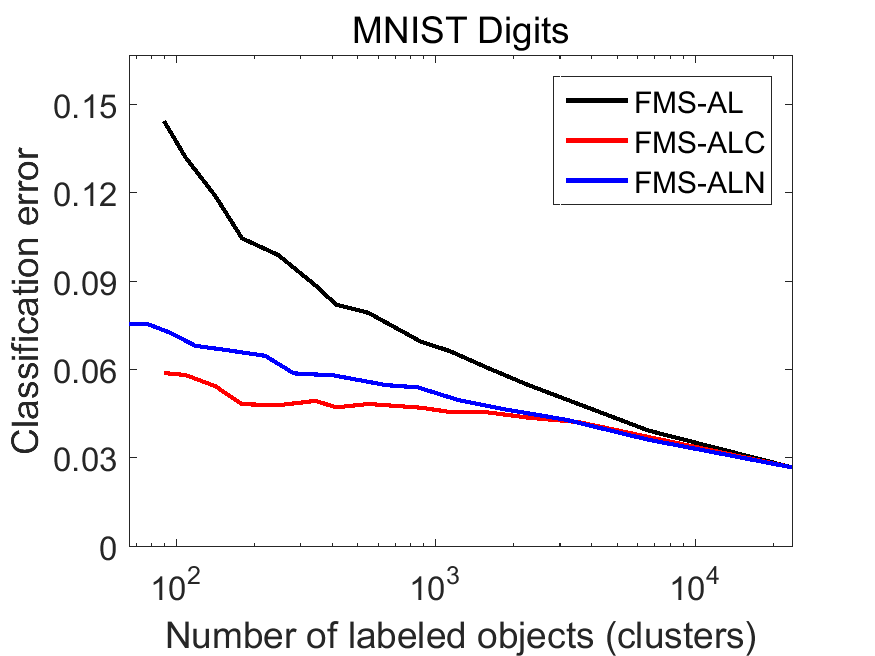} ~~
\includegraphics[width=6.5cm]{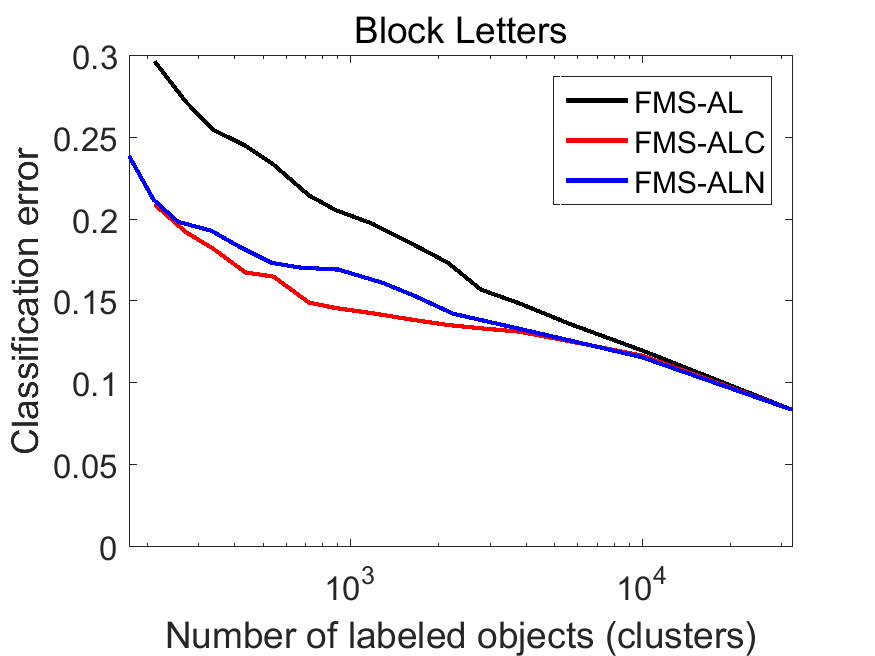}\\
\vspace{5mm}
\includegraphics[width=6.5cm]{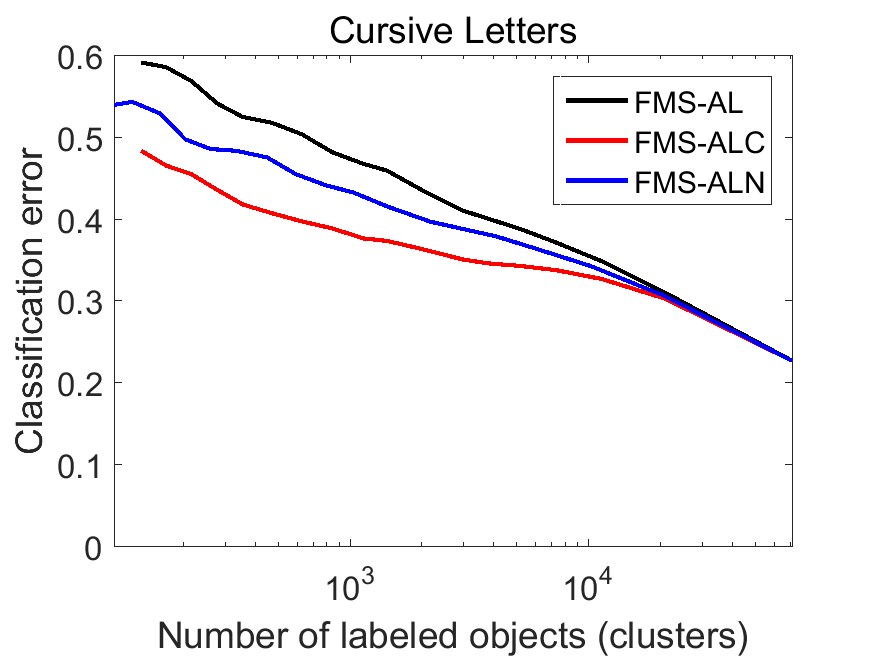}~~
\includegraphics[width=6.5cm]{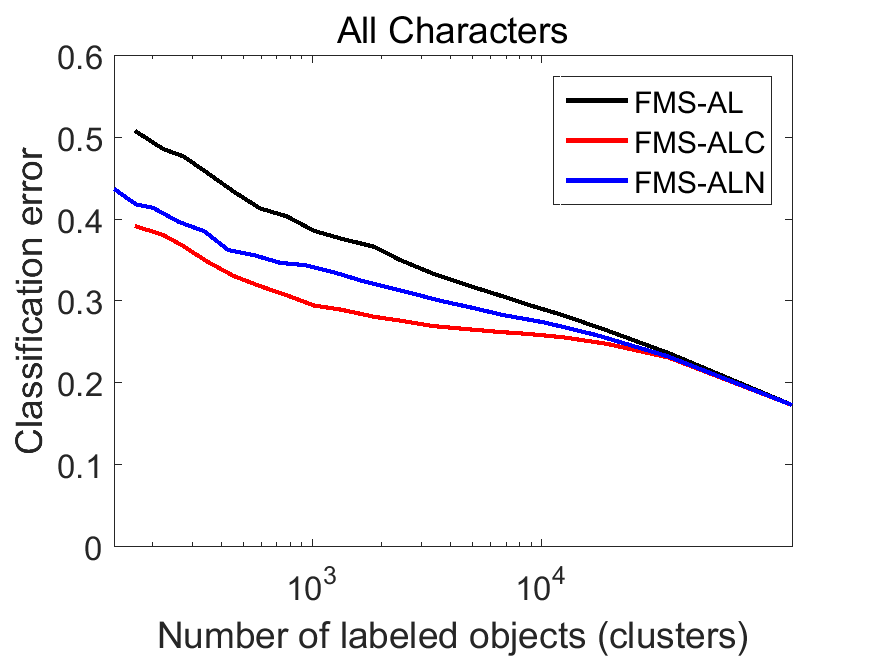}
\caption{Active Learning evaluation comparing active learning classification results based on a single clustering level, FMS-AL, with the combining the cluster level classifications, FMS-ALC, and with nesting, FMS-ALN.}
\vspace{-3mm}
\label{fig:ActLearning_Evaluation Combined}
\end{figure}

Next we studied the active learning classification results of combining all clustering levels as proposed in Section \ref{section:multi-level_clustering}. In Figure \ref{fig:ActLearning_Evaluation Combined} the original learning curves of MS-AL are compared with the combined classifiers, FMS-ALC, and and with nesting, FMS-ALN. It shows that significant improvements could be reached. The FMS-ALC procedure, however, is slow as for all objects the confidences have to be propagated and averaged over the clustering levels. Nesting is fast as it is based on changing cluster indices only.

%
%\begin{figure}[!h]
%\center
%\includegraphics[width=6.5cm]{LCurve5_Block} ~~
%\includegraphics[width=6.5cm]{LCurve5_MNIST}\\
%\vspace{5mm}
%\includegraphics[width=6.5cm]{LCurve5_Cursive} ~~
%\includegraphics[width=6.5cm]{LCurve5_All}\\
%\vspace{5mm}
%\includegraphics[width=6.5cm]{LCurve5_Allr}
%\caption{Clustering classification evaluation all datasets. Clustering is based on the the FMS-6 procedure. As a reference the 1-NN classification results are shown. They are based on the average of 25 repeats of randomly selected training sets.}
%\vspace{-3mm}
%\label{fig:LCurves}
%\end{figure}

\subsection{Reject curves}
\label{sec:reject_curves}

\begin{figure}[!h]
\center
\includegraphics[width=6.5cm]{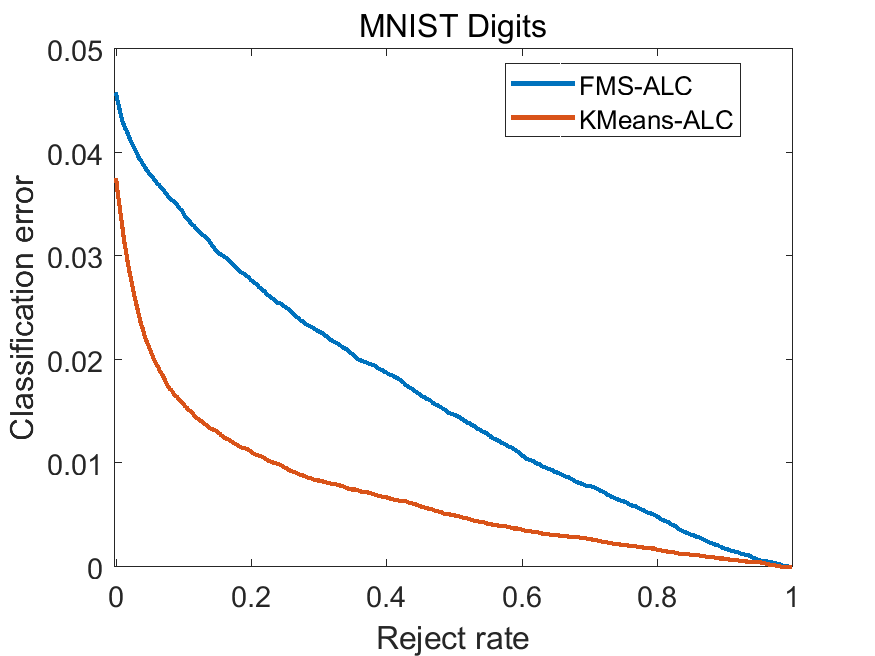} ~~
\includegraphics[width=6.5cm]{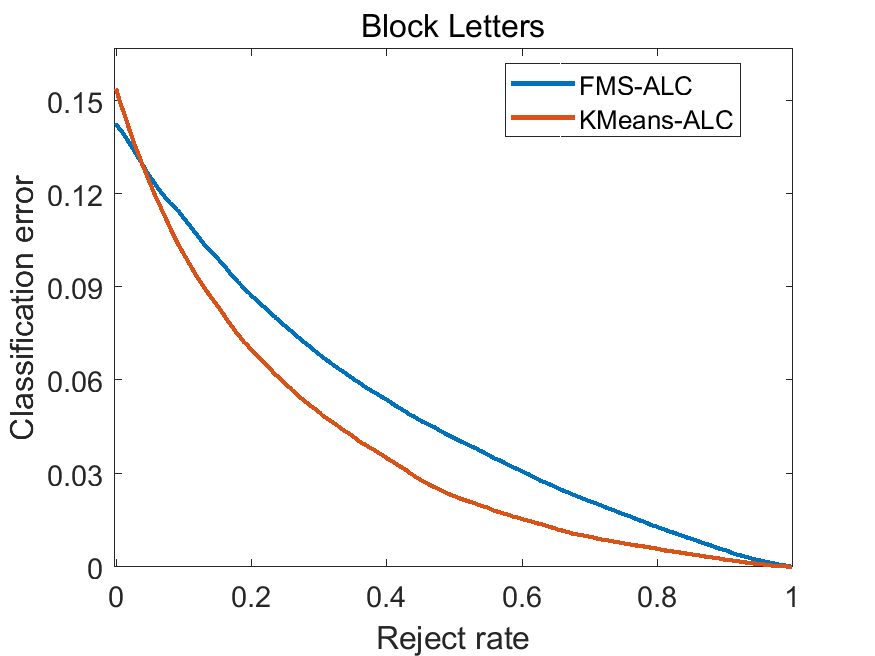}\\
\vspace{5mm}
\includegraphics[width=6.5cm]{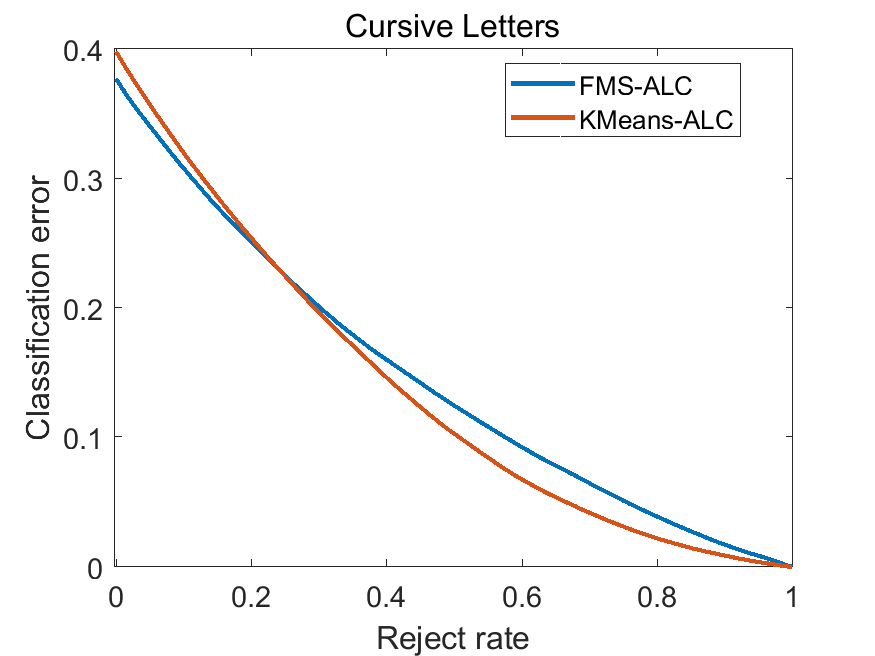} ~~
\includegraphics[width=6.5cm]{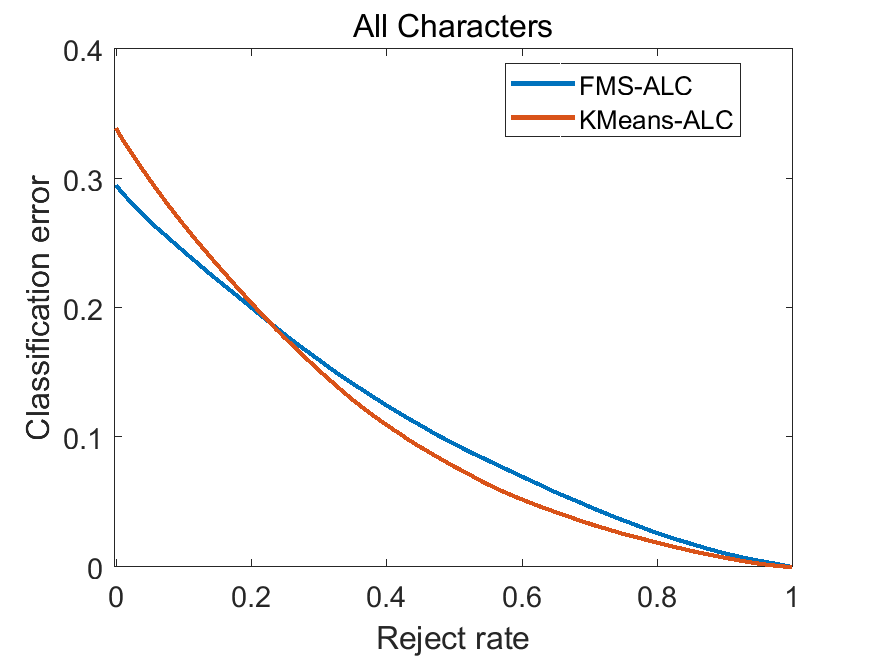}\\
\caption{Reject curves for the three datasets based on slightly more than 1000 labeled objects and confidences computed by FMS-ALC.}
\vspace{-3mm}
\label{fig:ActLearning_Evaluation Reject}
\end{figure}

The FMS-ALC and kMeans-ALC classifiers yield confidences that may be used for studying reject curves. As an example we considered cases of about 1000 clusters found by FMS, thereby generating training sets of bout 1000 objects. All other objects were classified and results with confidences lower than some threshold were rejected. The classification performances of the remaining objects (with high classification confidences) as a function of the rejects rate are shown in Figure \ref{fig:ActLearning_Evaluation Reject}.

The kMeans-ALC classifier shows better results. Its reject curve for the MNIST dataset is really good. It starts with rejecting just erroneously classified objects as for low rejects rates the error curve drops as fast as the reject rate goes up.

%\subsection{Semi-supervised Learning}
%\label{sec:ssl_experiments}
%
%In the semi-supervised learning (SSL) procedures the labeled objects are selected at random instead of by the modes of the density function. In our experiments this is repeated 5 times and averaged results are presented in Figure \ref{fig:LCurves}. It shows that the learning curves of the FMS-SSL are clearly worse than those of FMS-ALC. It can thereby be concluded that the modes are informative.
%
%For small sets of labeled objects some clusters of the initial layer (which has the smallest number of clusters), and possibly even some classes, have no training objects. Performances are deteriorated by this. For the initial layer we choose the one with less clusters than half the number of labeled objects. For large training sets, however, there is just a single layer that fulfils this requirements. Combining a set of layers is essential for the procedure. We therefor decided that at least five clustering layers should be used. This improves the use of a single layer, but it is still not good. Consequently, the curves show that the SSL results by using a large labeled training set are deteriorated by adding a small unlabeled set. This holds for many SSL procedures published in the literature.
%
%In conclusion, SSL may be useful for sufficiently large unlabeled sets, typically an order larger than the labeled set.  Moreover, the labeled set should at least include objects of all classes, preferably several ones to cover multi-modality of the data.

\newpage
\section{Conclusions}
\label{section:Conclusions}
\vspace{-1mm}
The main results of our study are:
\begin{itemize}
  \item We found a feasible way to find density information of datasets of millions of objects in high-dimensional spaces. This information is expressed in modal objects and pointers to them for every object in the dataset.
  \item The obtained density information is multi-scale as it is based on several neighborhood sizes simultaneously.
  \item The modal objects can be used to obtain a multi-level clustering, possible thousands of clusters in millions of objects.
  \item By labeling the modal objects a good active learning classifier is constructed that is better than classifiers trained by randomly selected subsets of the dataset.
  \item By combining the classification results of various clustering levels, classification confidences resulting in better classifiers are obtained.
  \item The time-complexity of the procedure is $O(n \sqrt n)$ resulting in seconds for $10^4$ objects to less than an hour for $10^6$ objects.
\end{itemize}

\section*{Acknowledgments} The authors thank Marco Loog, David Tax, Marcel Reinders and the members of the Delft Pattern Recognition Laboratory for support and hospitality. Prime Vision is acknowledged for supplying the datasets Block Letters and Cursive Letters.
{\small
\bibliography{knn_mode_seeking_ref}}

\begin{thebibliography}{1}
\providecommand{\url}[1]{\texttt{#1}}
\providecommand{\urlprefix}{URL }

\bibitem{Comaniciu02}
Comaniciu, D., Meer, P.: Mean shift: {A} robust approach toward feature space
  analysis. IEEE Trans. Pattern Anal. Mach. Intell  24(5),  603--619 (2002)

\bibitem{Dong2011}
Dong, W., Moses, C., Li, K.: Efficient k-nearest neighbor graph construction
  for generic similarity measures. In: Proceedings of the 20th international
  conference on World wide web. pp. 577--586. ACM (2011)

\bibitem{Duin2012}
Duin, R.P.W., Fred, A.L.N., Loog, M., Pekalska, E.: Mode seeking clustering by
  {KNN} and mean shift evaluated. In: SSPR/SPR. Lecture Notes in Computer
  Science, vol. 7626, pp. 51--59. Springer (2012)

\bibitem{Duin_ct}
Duin, R.: {C}luster{T}ools, a {M}atlab toolbox for cluster analysis and active
  labeling, \url{http://37steps.com/clustertools}

\bibitem{Fukunaga75}
Fukunaga, K., Hostetler, L.D.: The estimation of the gradient of a density
  function, with applications in pattern recognition. IEEE Trans. Information
  Theory  21(1),  32--40 (Jan 1975)

\bibitem{Jain10}
Jain, A.K.: Data clustering: 50 years beyond k-means. Pattern Recognition
  Letters  31(8),  651 -- 666 (2010)

\bibitem{Kaufman90}
Kaufman, L., Rousseeuw, P.J.: Finding Groups in Data: an Introduction to
  Cluster Analysis. Wiley (1990)

\bibitem{Kittler1998}
Kittler, J., Hatef, M., Duin, R.P.W., Matas, J.: On combining classifiers. IEEE
  Trans. Pattern Anal. Mach. Intell  20(3),  226--239 (1998)

\bibitem{Koontz76}
Koontz, W.L.G., Narendra, P.M., Fukunaga, K.: A graph-theoretic approach to
  nonparametric cluster analysis. IEEE Trans. Computer  25,  936--944 (1976)

\end{thebibliography}
\end{document}